\documentclass{article}

\usepackage[final]{neurips_2023}

\usepackage[utf8]{inputenc} 
\usepackage[T1]{fontenc}    
\usepackage{hyperref}       
\usepackage{url}            
\usepackage{booktabs}       
\usepackage{amsfonts}       
\usepackage{amssymb}
\usepackage{amsmath}
\usepackage{nicefrac}       
\usepackage{microtype}      
\usepackage{xcolor}         
\usepackage{graphicx}
\usepackage[ruled,vlined]{algorithm2e}
\usepackage{algorithmic}

\title{The Hidden Linear Structure in \\
Score-Based Models and its Application}

\author{%
  Binxu Wang\thanks{They are currently affiliated with The Kempner Institute for the Study of Natural and Artificial Intelligence at Harvard University.} \\
  Department of Neurobiology\\
  Harvard Medical School\\
  Boston, MA\\
  \texttt{binxu\_wang@hms.harvard.edu} \\
  \And
  John J. Vastola \\
  Department of Neurobiology\\
  Harvard Medical School\\
  Boston, MA\\
  \texttt{John\_Vastola@hms.harvard.edu} \\
}

\begin{document}

\maketitle

\begin{abstract}
  Score-based models have achieved remarkable results in the generative modeling of many domains. By learning the gradient of smoothed data distribution, they can iteratively generate samples from complex distribution e.g. natural images. 
  However, is there any universal structure in the gradient field that will eventually be learned by any neural network? Here, we aim to find such structures through a normative analysis of the score function. 
  First, we derived the closed-form solution to the scored-based model with a Gaussian score. We claimed that for well-trained diffusion models, the learned score at a high noise scale is well approximated by the linear score of Gaussian. We demonstrated this through empirical validation of pre-trained images diffusion model and theoretical analysis of the score function. This finding enabled us to precisely predict the initial diffusion trajectory using the analytical solution and to accelerate image sampling by 15-30\% by skipping the initial phase without sacrificing image quality. Our finding of the linear structure in the score-based model has implications for better model design and data pre-processing. 
\end{abstract}
\vspace{-7pt}
\section{Motivation: Universal structures of score}
The diffusion model or score-based model learns the gradient of the smoothed data distribution (i.e. \textit{score}), which is used to guide data sampling. Mathematically, the score function is a dynamic vector field in the domain of data, which changes with respect to time or noise scale in the diffusion process. Obviously, the score function depends on the points in the dataset, that's why a neural network is trained to approximate it. However, governed by the diffusion process, is there any universal structure of the score field that we can leverage? 

In this paper, we claimed that for diffusion models, the early phase of the score vector field is dominated by a linear structure, i.e. the score of Gaussian approximation of the data. We derived the closed-form solution to the Gaussian score model. We found that by leveraging this linear structure and the solution of diffusion trajectory, we can predict the early phase of the diffusion sampling trajectory and accelerate diffusion sampling by skipping this phase entirely, without affecting image quality. This finding has implications for how to train or design diffusion models. 

\section{Formulation: Score-based model}
We will use the unifying framework and notations in \cite{karras2022elucidatingDesignSp} throughout this paper. The score function $\nabla \log p(\mathbf{x},\sigma)$ is defined as the gradient of the log data distribution convolved by a Gaussian $\mathcal N(0,\sigma^2I)$. Here, we consider the deterministic sampler, i.e. sampling by integrating the probability flow Ordinary Differential Equation (ODE), with $\sigma_t$ a decreasing function of time $t$. 
\begin{align}\label{eq:probflow_ode_edm}
    d\mathbf{x} = -\dot{\sigma}_t\sigma_t\nabla_\mathbf{x} \log p(\mathbf{x},\sigma_t) dt
\end{align}
The score function is learned by minimizing the denoising score-matching objective in Eq. \ref{eq:denoisScoreMatch}. Following \cite{karras2022elucidatingDesignSp}, we reparametrize the score with the optimal `denoiser' $D_\theta$ using a neural network.  
\begin{align}\label{eq:denoisScoreMatch} 
    \mathcal{L}_\theta=\mathbb{E}_{\mathbf{y}\sim p_{data}}\mathbb{E}_{\mathbf{n}\sim\mathcal{N}(0,\sigma^2 I)}\|D_\theta(\mathbf{y}+\mathbf{n},\sigma)-\mathbf{y}\|^2_2.\;\Rightarrow \nabla_\mathbf{x} \log p(\mathbf{x},\sigma) \approx(D_\theta(\mathbf{x},\sigma) - \mathbf{x})/\sigma^2
\end{align}
Here, we will present results without time-dependent data scaling, but with little modification, all the results applied to models with data scaling e.g. DDPM \citep{ho2020DDPM}, DDIM \citep{song2020DDIM}, VP-ODE \citep{song2021scorebased}.  
 
\begin{figure*}[t]
\vspace{-16pt}
\centering
\includegraphics[width=1.0\textwidth]{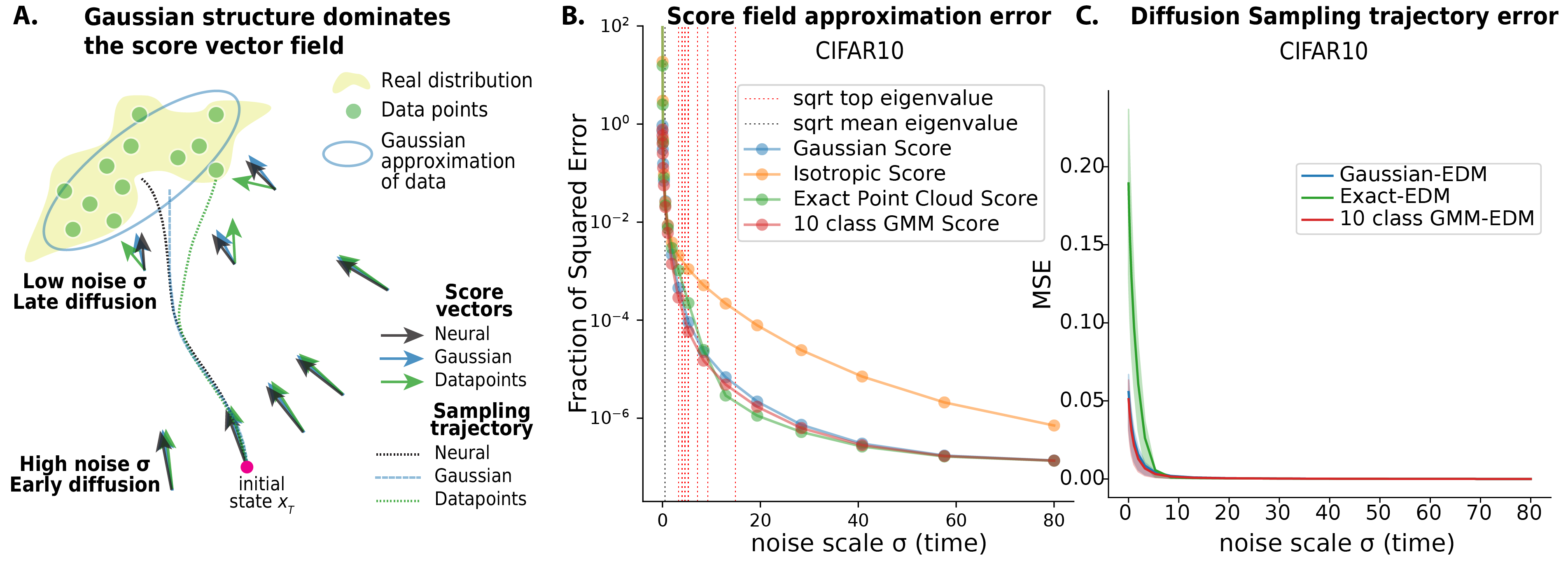}\vspace{-6pt}

\caption{\textbf{Gaussian score function approximates the learned neural score} \textbf{A.} Conceptual plan of the paper. \textbf{B.} Approximation error of learned neural score by different analytical scores: Gaussian, isotropic, Gaussian mixture, and `exact' point cloud. Note the log scale of error on y axis. \textbf{C.} Deviation of diffusion trajectory from the solution of analytical scores.}
\vspace{-10pt}
\label{fig:score_valid_cmp}
\end{figure*}
\section{Theory: Closed-form solution to diffusion with Gaussian score}


First, we showed that the diffusion trajectory is analytically solvable for Gaussian training distribution. Let $\mathbf{x}_t \in \mathbb{R}^D$, and the mean and covariance of the Gaussian be $\mathbf{\mu}$ and $\mathbf{\Sigma}$. Assuming $\mathbf{\Sigma}$ has rank $r \leq D$, it has a compact singular value decomposition (SVD) $\mathbf{\Sigma} = \mathbf{U} \mathbf{\Lambda} \mathbf{U}^T$, where $\mathbf{U} = [\mathbf{u}_1, ..., \mathbf{u}_r]$ is a $D \times r$ semi-orthogonal matrix. The columns of $\mathbf{U}$ are the principal axes along which the mode varies, and their span comprises the `data manifold'. 
At noise scale $\sigma$, the score is that of $\mathcal{N}(\mu,\Sigma+\sigma^2 I)= (\sigma^2 I+\Sigma)^{-1} (\mathbf{\mu} - \mathbf{x})$. With Woodbury matrix inversion trick \citep{woodbury1950inverting}, it simplifies as, 
\begin{align}\label{eq:gauss_score}
    \nabla \log p(\mathbf{x};\sigma)  = \frac{1}{\sigma^2}(I- \mathbf{U}\tilde \Lambda_\sigma  \mathbf{U}^T) (\mathbf{\mu}-\mathbf{x})\;,\;\;   \tilde \Lambda_\sigma &= \mbox{diag}\big[\frac{\lambda_k}{\lambda_k+\sigma^2}\big]
\end{align}
which is a linear function of state. It consists of an isotropic term pointing to the mean $\mathbf{\mu}$, and a 2nd term depending on the PC spectrum structure $\mathbf{U},\mathbf{\Lambda}$. 
With this linear score function, probability flow (Eq. \ref{eq:probflow_ode_edm}) becomes a linear ODE, 
which has the following closed-form solution (Appendix \ref{apd:deriv_gaussian_model}). 
\begin{align}  \label{eq:xt_solu_psi_def}
    \mathbf{x}_t = &\mathbf{\mu} + \frac{\sigma_t}{\sigma_T} \ \mathbf{x}^{\perp}_T + \sum_{k=1}^r \psi(t, \lambda_k) c_k(T) \mathbf{u}_k  \;\; 
    &\psi(t, \lambda_k):=\sqrt{  \frac{\sigma_t^2 + \lambda_k}{\sigma_T^2 + \lambda_k} }  \\ 
    \mathbf{x}^{\perp}_T:=&\ (\mathbf{I}-\mathbf{U}\mathbf{U}^T)(\mathbf{x}_T- \mathbf{\mu}) \; 
    &c_k(T) := \ \mathbf{u}_k^T(\mathbf{x}_T-\mathbf{\mu}) \ .
\end{align}  
The solution has 3 components, corresponding to the 3 parts of the initial condition $\mathbf{x}_T$: 1) the distribution mean $\mathbf \mu$ stays; 2) the off-manifold component $\mathbf{x}^{\perp}$ shrinks proportional to the noise scale $\psi(t,0)=\frac{\sigma_t}{\sigma_T}$; and 3) the projections along each eigenvector $\mathbf{u}_k$ moves governed by $\psi(t,\lambda_k)$. 
Similarly, the ideal denoising image along the trajectory can be expressed as 
\begin{align}\label{eq:denoiser_gauss_solu}
    D(\mathbf{x}_t,\sigma_t)= \  \mathbf{\mu} + \sum_{k=1}^r 
\xi(t,\lambda_k) c_k(T) \mathbf{u}_k \hspace{0.5in}
\xi(t,\lambda):=\frac{\lambda}{\sqrt{(\lambda + \sigma_t^2)(\lambda + \sigma_T^2)}} \ .
\end{align}
Note that even this simple closed-form solution can explain many phenomena of diffusion models, including 1) the sampling trajectory of variance-preserving diffusion models tend to be low dimensional, and more specifically, resembling a two-dimensional rotation on the plane of initial and final state; 2) the outlines or low-frequency aspect of images are determined earlier in the generation process; 3) earlier and on manifold perturbations tend to have a larger impact on the image generation process. 
Though not directly related to the main claim of this paper, we encourage the readers to further explore these connections in \cite{wang2023diffusionpainter}. This showed that qualitatively, the sampling dynamics of diffusion models are largely dominated by the Gaussian solution. 

\begin{figure*}[!ht]
\vspace{-3pt}
\centering
\includegraphics[width=1.0\textwidth]{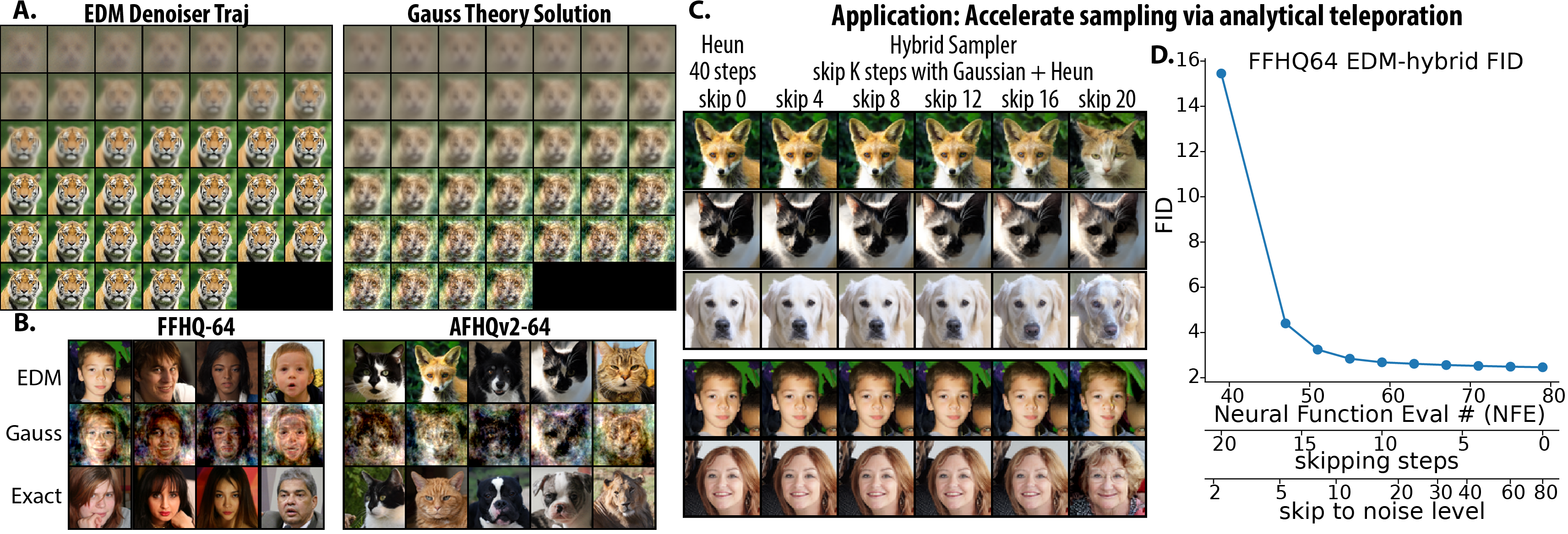}\vspace{-4pt}
\caption{\textbf{Validation and application of the Gaussian score approximation}. 
\textbf{A.} The denoiser images $D(\mathbf{x}_t,t)$ along a diffusion sampling trajectory compared with the Gaussian solution with the same initial condition $\mathbf{x}_T$. 
\textbf{B.} Samples generated by the EDM model, Gaussian solution, and the `exact' scores from the same initial condition. 
\textbf{C.} Sampled image as a function of skipped steps with the hybrid method combining Gaussian theory with Heun's method. 
\textbf{D.} Image quality (FID score) of the hybrid method as a function of NFE, skipped steps, and skipped noise scale (see Appendix \ref{apd:fid_method}).}
\label{fig:EDM_theory_valid}
\vspace{-9pt}
\end{figure*}

\section{Validation: The Gaussian score structure in diffusion models}
Most distributions useful enough to be fit by generative models are not Gaussian. So how is the analytical solution of the Gaussian score model related to practical diffusion models? 
Here we empirically tested our claim: for pre-trained image diffusion models, the score in the early high-noise phase is dominated by the linear score of the Gaussian approximation of data. 
In Appendix \ref{apd:gauss_pointcloud_equiv}, we also argued theoretically that, at a high enough noise scale, the score function of bounded point cloud is indistinguishable from that of the Gaussian with matching mean and covariance. 

We picked several image datasets (CIFAR10, FFHQ64, AFHQv2-64 \citep{choi2020starganv2}) and the diffusion models pre-trained on them using the optimal configuration in \cite{karras2022elucidatingDesignSp}. 
For each dataset, we computed several analytical approximations of the score and compared that with the score function of the neural network (Eq. \ref{eq:denoisScoreMatch}): 1) Isotropic score (\textit{Iso}), $(\mathbf{\mu} - \mathbf{x})/\sigma^2$, which is the first order term of the Gaussian score; 2) Gaussian score (Eq. \ref{eq:gauss_score}), computed with the mean and covariance of the training dataset; 3) Point cloud score (\textit{Exact}), computed for Gaussian mixture model with a Gaussian mode with negligible width at each data point; 4) Per-class Gaussian Mixture Score (\textit{GMM}), when the dataset has class labels (e.g. CIFAR), we computed the score for Gaussian Mixture Model where each mode has mean and covariance for each class (details in Sec.\ref{apd:ideal_score_eqs}). 
We compared these score vector fields with the neural scores, and the diffusion trajectories guided by them. 


\paragraph{Gaussian score predicts the learned score at high noise} 
First, we validated the score vector prediction. For each noise scale $\sigma$, we evaluated the neural and analytical scores at random points sampled from $\mathcal N(0, \sigma^2 I)$. 
We computed the fraction of unexplained variance (residual) as a function of the noise scale for each analytical score approximator (Fig. \ref{fig:score_valid_full}). For all three spaces, at most noise levels ($\sigma > 1.08$), the Gaussian score explains almost all variance (> 99\%) of the neural score. 
As a control, at all noise levels, the Gaussian score explains more variance than the isotropic score without information of the covariance. 
Further, for CIFAR10, at all levels, the 10-class Gaussian mixture score predicts neural score slightly better than the Gaussian score, which shows that adding more modes is indeed helpful to capture the details of the score field. 

\paragraph{Neural score deviates from `exact' score at low noise} However, this trend breaks down when taken to an extreme. 
In all three datasets, the exact point cloud score shows a slight improvement over the neural score in high-noise conditions (Fig. \ref{fig:score_valid_full}); however, it diverges significantly from the neural score under low-noise conditions.
This dichotomy implies that in high-noise regimes, the neural score captures more nuanced features of the data points than a single Gaussian does. Conversely, in low-noise regimes, the neural score diverges from the `exact' score, resembling more closely a Gaussian or Gaussian mixture score.
This indicates that the neural network learned a smoother score field, particularly in low-noise conditions \citep{kadkhodaie2023generalization,scarvelis2023closed}, thereby mitigating overfitting and preventing memorization, since learning the exact score would lead to generating samples exclusively from the training set. 

 
\paragraph{Gaussian solution predicts early diffusion trajectory}
Next, we tested how well the solutions to ODE (Eq. \ref{eq:probflow_ode_edm}) with the closed-form scores predicted the sampling trajectories of actual diffusion models. 
We found that the early phase of reverse diffusion is well-predicted by the Gaussian solution (Fig. \ref{fig:valid_xt_traj}, \ref{fig:valid_denoiser_traj}). 
The predicted and actual trajectory stayed close (MSE < 0.01) for around 9 sampling steps (noise scale $\sigma_t=1.92$) for CIFAR10 and 17 steps ($\sigma_t=4.37$) for FFHQ and AFHQ until diverging. This roughly matches the scale where the Gaussian approximation of the score field breaks down, and the score needs to be approximated by that of a more complicated distribution (Fig.\ref{fig:score_valid_full}). 
Visually, as the low-frequency information is determined early on \cite{ho2020DDPM}, the `layout' or shading of the final image is predicted by Gaussian solution, while high-frequency details such as edges are less well-predicted (Fig. \ref{fig:EDM_theory_valid} A,B). 

\paragraph{Neural diffusion trajectory deviates from the `exact' score at low noise} 
In line with our score field observations, the early diffusion trajectory aligns better with the exact score trajectory than with the Gaussian score trajectory. However, the late trajectory based on neural score diverges from the exact score trajectory, yielding notably different samples in the end (Fig. \ref{fig:EDM_theory_valid} B).

In summary, these results show that at the high noise regime, the neural score function is dominated by the Gaussian approximation, but with details better predicted by the more elaborate Gaussian mixture score or `exact' score. But at low noise regime, the actual neural score function learned is different from the `exact' score, and it's more similar to the `smoother' score of Gaussian or Gaussian mixture, possibly due to the regularizing effect of the neural network function approximator. 

\section{Application: Accelerating Diffusion with analytical teleportation.}
\label{sec:app_teleportation}
The Gaussian analytical solution provides a surprisingly good approximation to the early sampling trajectory, thus a direct application is to accelerate diffusion by `teleportation'. 
Namely, instead of evaluating the neural score function and integrating the probability flow ODE, we can directly evaluate $\mathbf{x}_t$ using the Gaussian solution with $\mathbf{\mu}$ and $\mathbf{\Sigma}$ of the training set. In principle, this speedup can be combined with any deterministic sampler. We showcase its effectiveness with the optimized 2nd order Heun sampler from \cite{karras2022elucidatingDesignSp}, which has demonstrated near state-of-the-art image quality and efficiency. 
For each initial state, we skipped to different times with Eq.\ref{eq:xt_solu_psi_def} and continued with the Heun sampler from that time (see Sec.\ref{apd:hybrid_algor_detail}, Alg.\ref{algr:hybrid}). 
We tested this hybrid sampler on unconditional diffusion models of CIFAR-10, FFHQ-64, and AFHQv2-64 \cite{karras2022elucidatingDesignSp}. 
We found that, consistently, we can save 15-30\% of neural function evaluation (NFE) with less than 3\% increase in the FID score (Fig.\ref{fig:EDM_theory_valid} C,D). Intriguingly, for CIFAR10 and AFHQ, skipping steps can even reduce the FID score by 1\% (full results in Sec.\ref{apd:fid_ext_result} Fig.\ref{fig:hybrid_fid_fullresult},  Tab.\ref{tab:ffhq64_fid},\ref{tab:afhqv264_fid},\ref{tab:cifar10_fid}). We also noticed that the number of skippable steps roughly corresponds to the noise scale where the neural score deviates from the Gaussian proxy. 

\section{Discussion}
In summary, we showed that even for natural images, the learned score at a high noise regime is well predicted by the linear score of Gaussian; at the low noise regime, it is better approximated by a smoother Gaussian (Mixture) score instead of an `exact' score. We leveraged this finding to accelerate diffusion sampling. This result bears further implications for training and designs of diffusion models. 


\textbf{Implications on Model Design} Since the score function to be learned is dominated by the Gaussian score, we can 
we can directly build this linear structure into the neural network to assist learning, e.g. to add a by-pass pathway based on the Gaussian approximation score of training distribution, and let the neural network learn the nonlinear residual to the Gaussian terms only. 

\textbf{Implications on Training Distribution} 
Further, we may be able to reshape the training distribution to assist learning. 
We hypothesize that if we pre-condition the target distribution by whitening its spectrum, then the target score become more isotropic, and the score network may converge faster. 
This might explain the training efficiency of latent diffusion models \cite{rombach2022latentdiff}: with the KL regularization, the auto-encoder not only compressed the state space but also `whiten' the image distribution by morphing it closer to isotropic Gaussian distribution. Similar whitening tricks is worth exploring for general diffusion models. 

\begin{ack}
B.W. was funded by Victoria Quan fellowship at Harvard Medical School, and Kempner research fellowship. 
We thank Carlos R. Ponce lab, and O2 Cluster at Harvard Medical School for providing computational resources for this work. 
We thank Zhengdao Chen, Jacob Zavatone-Veth, Cengiz Pehlevan, Yao Fu, Hao Sun for providing feedback to an early version of this work. 
\end{ack}
\bibliographystyle{apalike}
\bibliography{neurips_diff}

\clearpage
\section{Detailed Methods}
\subsection{Image datasets and Models}

\begin{table}[!h]
    \centering
    \caption{\textbf{Specifications of the image datasets and diffusion models}}
    \begin{tabular}{r|ccc}
    \toprule
        Dataset name & Num Samples & Resolution & Pre-trained Model Spec\\
        \midrule
        CIFAR10 & 50000 & 32 & \texttt{edm-cifar10-32x32-uncond-vp} \\
        FFHQ64  & 70000 & 64 & \texttt{edm-ffhq-64x64-uncond-vp}\\
        AFHQv2-64 & 15803 & 64 & \texttt{edm-afhqv2-64x64-uncond-vp} \\
        \bottomrule
    \end{tabular}
    \label{tab:dataset}
\end{table}

\subsection{Idealized Score Approximations}\label{apd:ideal_score_eqs}
In the paper, we compared several analytical approximations of the score. We listed their formula below. 

\textbf{Isotropic score}, only depends on the mean of data, isotropically pointing towards $\mu$. 
\begin{equation}
    \frac{\mu-\mathbf{x}}{\sigma^2}
\end{equation}
This is equivalent to approximating the whole training dataset with its mean. 

\textbf{Gaussian score}, for a Gaussian distribution of $\mathcal N(\mathbf{x};\mathbf{\mu},\mathbf{\Sigma})$, its score is 
\begin{align}
& \nabla_{\mathbf{x}}\log\mathcal N(\mathbf{x};\mathbf{\mu},\sigma^2 \mathbf{I}+\mathbf{\Sigma}) \\
    &=(\sigma^2 I +\Sigma)^{-1}(\mu-\mathbf{x}) \\
    &= \frac{1}{\sigma^2}(I-U\tilde \Lambda_\sigma  U^T) (\mathbf{\mu}-\mathbf{x})\\
    &=\frac{\mu-\mathbf{x}}{\sigma^2} - \frac{1}{\sigma^2}U\tilde \Lambda_\sigma  U^T(\mathbf{\mu}-\mathbf{x})\\
    \tilde \Lambda_\sigma &= diag\big[\frac{\lambda_k}{\lambda_k+\sigma^2}\big]
\end{align}
it depends on the mean and covariance of the data. It added a non-isotropic correction term based on covariance to the isotropic score. 

\textbf{Gaussian mixture score}. For a general Gaussian mixture with $k$ component $q(\mathbf{x}) = \sum^k_i \pi_i \mathcal N(\mathbf{x};\mathbf{\mu}_i,\mathbf{\Sigma}_i)$, its score function at noise scale $\sigma$ is the following, 
\begin{equation}
\begin{split}
& \nabla_{\mathbf{x}}\log\Big(\sum^k_i \pi_i \mathcal N(\mathbf{x};\mathbf{\mu}_i,\sigma^2 \mathbf{I}+\mathbf{\Sigma}_i)\Big) \\
=&\sum_i -(\sigma^2 \mathbf{I}+ \mathbf{\Sigma}_i)^{-1}(\mathbf{x}- \mathbf{\mu}_i)\frac{\pi_i \mathcal N(\mathbf{x}; \mathbf{\mu}_i,\sigma^2 \mathbf{I}+\mathbf{\Sigma}_i)}{\sum^k_j \pi_j \mathcal N(\mathbf{x};\mathbf{\mu}_j,\sigma^2 \mathbf{I}+\mathbf{\Sigma}_j))}\\
=&\sum_i -(\sigma^2 \mathbf{I}+ \mathbf{\Sigma}_i)^{-1}(\mathbf{x}-\mathbf{\mu}_i)w_i(\mathbf{x}, \sigma) \ .
\end{split}
\end{equation}
Where the covariance matrix for each Gaussian component can be eigen-decomposed and inverted efficiently. It's a weighted average of the score of each Gaussian mode, with a softmax-like weighting function $w_i(\mathbf{x}, \sigma)$. 

\textbf{Exact point cloud score}. For a set of data points $\{\mathbf{y}_i\}$, the score of $\mathbf{x}$ at noise scale $\sigma$ is
\begin{align}
    & \nabla_{\mathbf{x}}\log\Big(\sum^N_i \frac{1}{N} \mathcal N(\mathbf{x};\mathbf{y}_i,\sigma^2 \mathbf{I}\Big)\\
    =&\frac{1}{\sigma^2} \left[- \mathbf{x} +  \sum_i w_i(\mathbf{x},\sigma)\mathbf{y}_i \right]\\
    =& \frac{1}{\sigma^2} \left[- \mathbf{x} +  \sum_i \frac{\exp\big(-\frac{1}{2\sigma^2}\|\mathbf{y}_i - \mathbf{x}\|^2\big)}{\sum_j\exp\big(-\frac{1}{2\sigma^2}\|\mathbf{y}_j - \mathbf{x}\|^2\big)} \mathbf{y}_i \right] \ .\\
    w_i(\mathbf{x},\sigma):=&\frac{\exp\big(-\frac{1}{2\sigma^2}\|\mathbf{y}_i - \mathbf{x}\|^2\big)}{\sum_j\exp\big(-\frac{1}{2\sigma^2}\|\mathbf{y}_j - \mathbf{x}\|^2\big)}=\mbox{softmax}\Big(-\frac{1}{2\sigma^2}\|\mathbf{y}_i - \mathbf{x}\|^2\Big)
\end{align}

\newpage
\subsection{Sampling Diffusion trajectory with Gaussian Mixture scores} 
For the Gaussian score, we are able to compute the whole sampling trajectory analytically. But for a Gaussian mixture with more than one mode, we need to use numerical integration. We evaluated the numerical score with the Gaussian mixture model defined as above and integrated Eq. \ref{eq:probflow_ode_edm} with off-the-shelf Runge-Kutta 4 integrator (\texttt{solve\_ivp} from \texttt{scipy}). We chose $\sigma(t)=t$, and integrate $t$ backward from sigma $80.0$ to $0.0$. To compare the trajectory with the one sampled with the original Heun method, we evaluated the trajectory at the same discrete time steps as the \cite{karras2022elucidatingDesignSp}
 paper. The $i$th noise level is the following, 
\begin{equation}\label{eq:edm_sigam_scheme}
     \sigma_i= \Big(\sigma_{max}^{1/\rho} + \frac{i}{n_{step} - 1}(\sigma_{min}^{1/\rho} - \sigma_{max}^{1/\rho})\Big)^{\rho}
\end{equation}
we chose the same hyper parameter $\sigma_{min},\sigma_{max},\rho$ and $n_{step}$ as the original EDM paper. 

\subsection{Hybrid sampling method}\label{apd:hybrid_algor_detail}
As we stated in the paper, the Hybrid sampling scheme can be combined with any deterministic sampler (e.g. DDIM \cite{song2020DDIM}, PNDM \cite{liu2022PNDM}). Here we detailed its implementation with Heun's method. 

\begin{algorithm}[H]\label{algr:hybrid}
\caption{Hybrid Sampling Method with Heun's method}
\begin{algorithmic}[1]
\REQUIRE Data: mean and covariance of dataset $ \mu $,  $ \Sigma $ and its eigendecomposition $\mathbf{\Sigma} = \mathbf{U} \mathbf{\Lambda} \mathbf{U}^T$
\REQUIRE Original parameter: $\sigma_{min},\sigma_{max},\rho$
\REQUIRE Parameter: Skip time/noise scale $ t'=\sigma_{skip} $
\REQUIRE Input: $\mathbf{x}_T$

    \STATE $ \mathbf{x}_{t'} \gets $ Gaussian solution at $ t' $ using $ \mu $ and $ \Sigma $: \\
    $\mathbf{x}_{t'} = \mathbf{\mu} + \frac{\sigma_{t'}}{\sigma_{max}} (\mathbf{I}-\mathbf{U}\mathbf{U}^T)(\mathbf{x}_T- \mathbf{\mu}) + \sum_{k=1}^r \sqrt{  \frac{\sigma_{t'}^2 + \lambda_k}{\sigma_{max}^2 + \lambda_k}} \mathbf{u}_k \mathbf{u}_k^T(\mathbf{x}_T-\mathbf{\mu})$
    \STATE Continue sampling using Heun method with $ \sigma_{\text{max}} = t' $ and initial state $\mathbf{x}_{t'}$, get the final sample $ \mathbf{x}_0 $. 
    \RETURN $ \mathbf{x}_0 $
\end{algorithmic}
\end{algorithm}

For a closer comparison with the baseline Heun method, we used the following strategy for choosing $\sigma_{skip}$. 
The Heun method will sample a sequence of $n_{step}$ noise levels $[\sigma_0,\sigma_1,\sigma_2,...\sigma_{n_{step}}]$, where $\sigma_0=\sigma_{max}$ and $\sigma_{n_{step}}=\sigma_{min}$. 
For each initial condition $\mathbf{x}_T$, we use the analytical solution to evaluate $\mathbf{x}_{t'}$ or integrate the probability flow ODE to time $t'$, where we chose $t'$ as the $i$-th noise level $\sigma_i$ (Eq. \ref{eq:edm_sigam_scheme}). Then we will skip the first $i$ step in the Heun method and start at initial state $\mathbf{x}_{t'}$. Note that Heun's method is a 2nd order method, which evaluates the neural score function twice for each step in $\mathbf{x}_t$. Because of this, skipping $i$ time steps will save $2i$ neural function evaluations (note x-axes in Fig. \ref{fig:hybrid_fid_fullresult}). 

\subsection{FID score computation and baseline} \label{apd:fid_method}
We used the same code for FID score computation as in \cite{karras2022elucidatingDesignSp}. For each sampler, we sampled the same initial noise state $x_T$ with random seeds 0-49999. We computed the FID score based on 50,000 samples. 

For our baseline, we picked the same model configurations as reported in Tab. 2, specifically, Variance preserving (VP) config F, the unconditional model for CIFAR10, FFHQ 64 and AFHQv2 64. The default sampling steps are 18 time steps ($NFE=35$) for CIFAR10, and 40 time steps ($NFE=79$) for FFHQ and AFHQ.

\clearpage
\section{Extended Results}
\subsection{Detailed Validation Results}
\begin{figure*}[!ht]
\vspace{-2pt}
\centering
\includegraphics[width=1.0\textwidth]{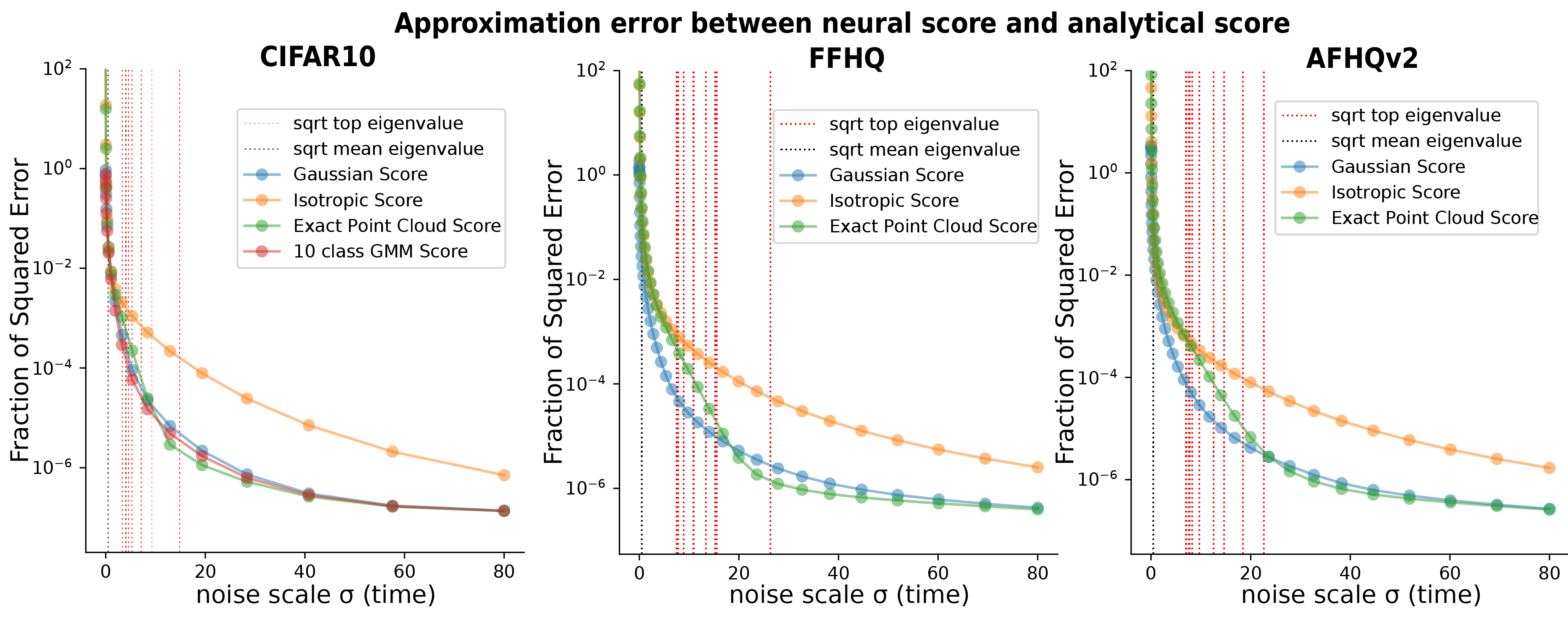}\vspace{2pt}
\includegraphics[width=1.0\textwidth]{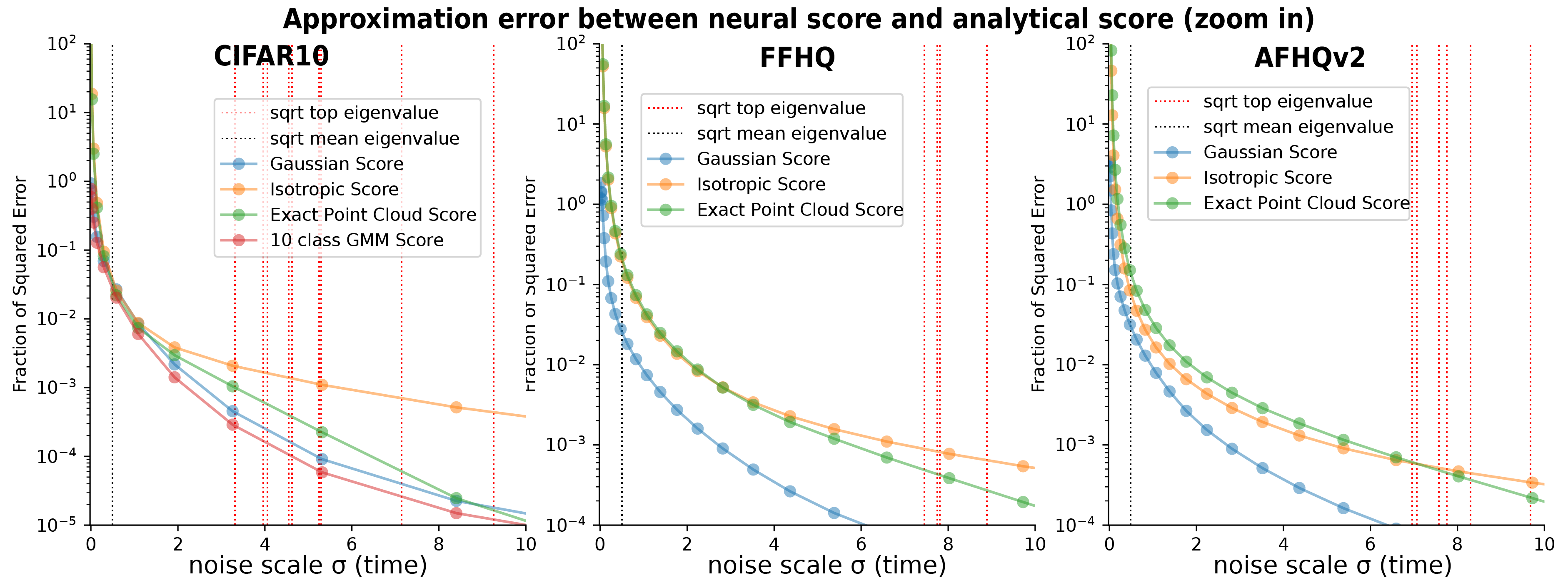}\vspace{-6pt}
\caption{\textbf{Approximation error of score learned by score neural network with various analytical approximations}. Note the log scale on y.}
\vspace{-2pt}
\label{fig:score_valid_full}
\end{figure*}

\begin{figure*}[!ht]
\vspace{-2pt}
\centering
\includegraphics[width=1.0\textwidth]{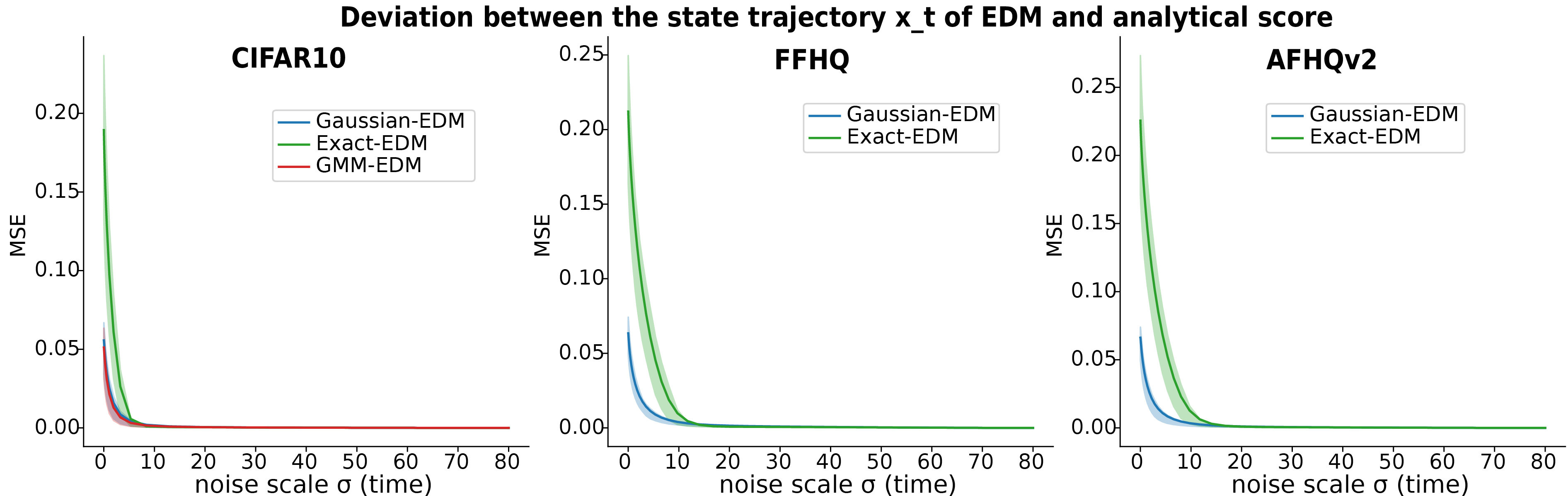}\vspace{-5pt}
\caption{\textbf{Deviation between the state trajectory $\mathbf{x}_t$ of EDM neural network and the analytical score}. The thick line denotes the mean over initial conditions; the shaded area denotes 25\%, 75\% quantile range over the initial conditions.}
\vspace{-2pt}
\label{fig:valid_xt_traj}
\end{figure*}

\begin{figure*}[!ht]
\vspace{-2pt}
\centering
\includegraphics[width=1.0\textwidth]{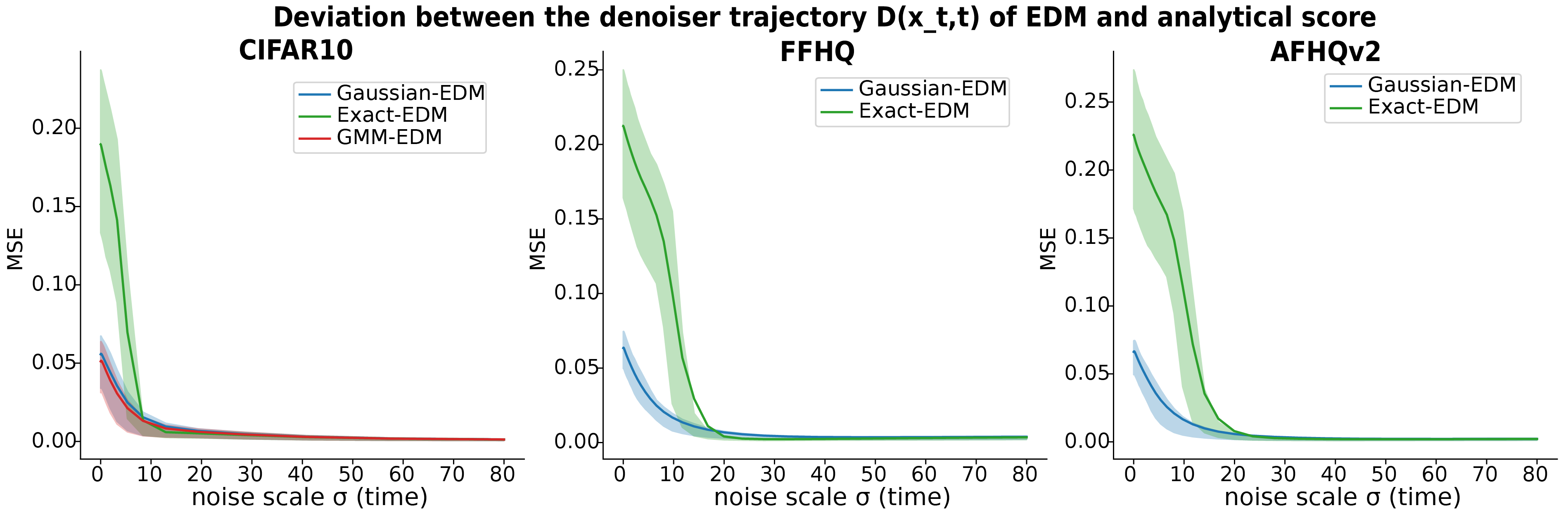}\vspace{-5pt}
\caption{\textbf{Deviation between the denoiser $D(\mathbf{x}_t,t)$ of EDM neural network and the analytical score}. The thick line denotes the mean over initial conditions; the shaded area denotes 25\%, 75\% quantile range over the initial conditions.}
\vspace{-2pt}
\label{fig:valid_denoiser_traj}
\end{figure*}

Here we presented the full results for all datasets: approximating neural scores with analytical scores (Fig.\ref{fig:score_valid_full}), deviations between sampling trajectory (Fig.\ref{fig:valid_xt_traj}) and denoiser trajectory (Fig.\ref{fig:valid_denoiser_traj}) guided by neural score and analytical score. 

\begin{figure*}[!ht]
\vspace{-2pt}
\centering
\includegraphics[width=1.0\textwidth]{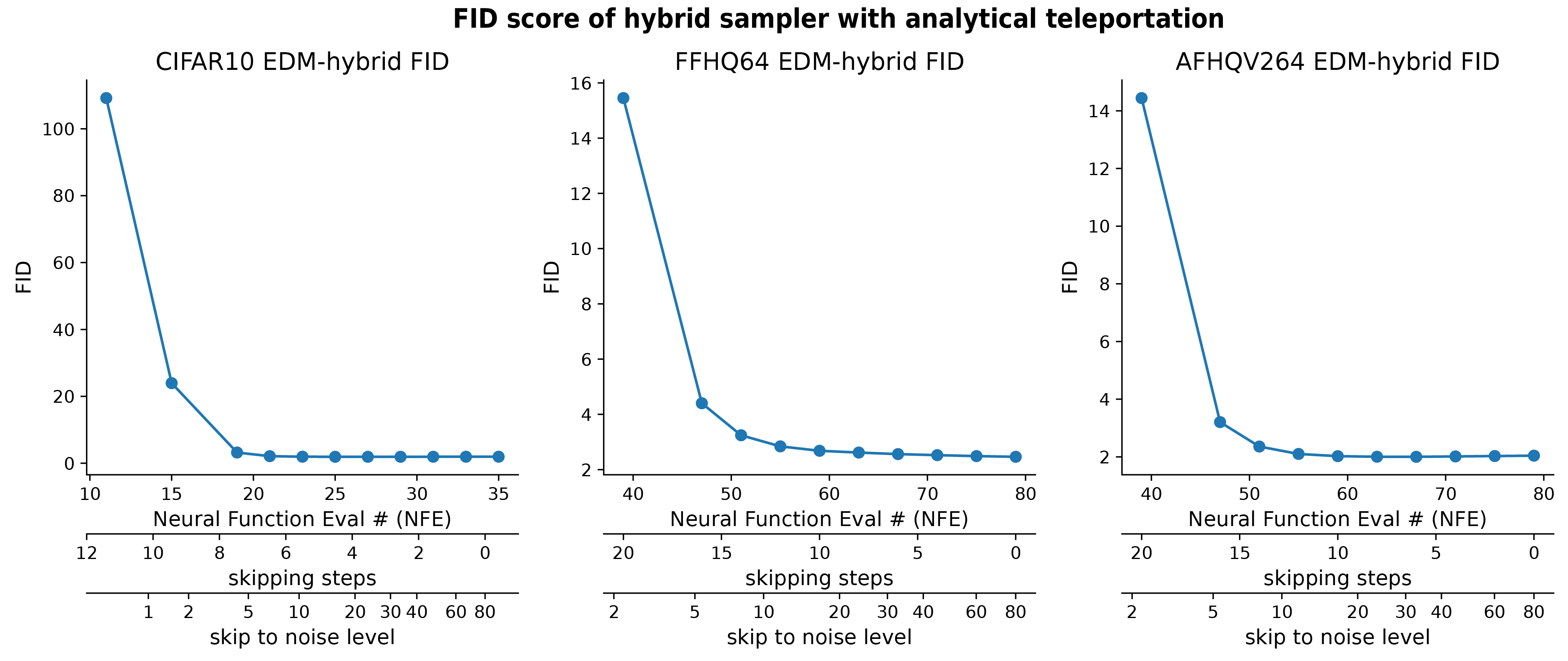}\vspace{-5pt}
\caption{\textbf{Image quality as a function of skipping steps for hybrid sampling approach.} Note the main x-axes are the number of Neural Function Evaluation (NFE); the secondary x-axes are the number of skipping steps from the Heun sampler; the tertiary-axes are the time or noise level $\sigma_{skip}$ at which we evaluate the Gaussian solution.  
See Tab.\ref{tab:cifar10_fid},\ref{tab:ffhq64_fid},\ref{tab:afhqv264_fid} for numbers.}
\vspace{-2pt}
\label{fig:hybrid_fid_fullresult}
\end{figure*}

\subsection{Detailed Gaussian teleportation FID scores results}\label{apd:fid_ext_result}
The effects of applying the Gaussian analytical teleportation on the FID scores are presented below (Fig.\ref{fig:hybrid_fid_fullresult}, Tab.\ref{tab:afhqv264_fid},\ref{tab:ffhq64_fid},\ref{tab:cifar10_fid}). 

For CIFAR-10 model, we found teleportation not only reduces the number of required neural function calls, but also lowers the FID score. Admittedly, the FID score effect is quite small on the CIFAR-10 model: it goes from 1.958 (no teleportation) to 1.933 (5 skipping steps), which is around 1\% decrease. 
Visually, the difference in sample images is mostly negligible; there are some changes, but they are plausible changes in image details. Hence, teleportation saves 29\% of NFE and improves the sample quality, without changing the model or the sampler. Further, skipping 6 steps will save 34\% NFE but increase the FID score by less than .2\%.

For AFHQv2-64, the teleportation can save 25-30\% of NFE (10-12 steps skipped) or reduce the FID by 2\% (6 steps skipped). For FFHQ64, the teleportation saves 10-15\% of NFE (4-6 steps skipped) with around 3\% increase in FID.

The full table of FID scores as a function of the number of skipped steps is shown below.

The fact that replacing neural network evaluations with the analytical solution can even improve the already low FID score (albeit very slightly in the tests we just ran) is intriguing, and it is not immediately obvious why this is true. One possibility is that, though using the Inception feature space, the FID score is still measuring the similarity of image distribution with a Gaussian approximation. Thus using an explicit Gaussian approximation of data to guide diffusion may help this metric. 

\begin{table}[!ht]
\centering
\caption{FFHQ64 FID with analytical teleportation}
\label{tab:ffhq64_fid}
\begin{tabular}{rrrr}
\toprule
 Nskip &   NFE &   time/noise scale &    FID \\
\midrule
     0 &    79 &             80.0 &  2.464 \\
     2 &    75 &             60.1 &  2.489 \\
     4 &    71 &             44.6 &  2.523 \\
     6 &    67 &             32.7 &  2.561 \\
     8 &    63 &             23.6 &  2.617 \\
    10 &    59 &             16.8 &  2.681 \\
    12 &    55 &             11.7 &  2.841 \\
    14 &    51 &              8.0 &  3.243 \\
    16 &    47 &              5.4 &  4.402 \\
    20 &    39 &              2.2 & 15.451 \\
\bottomrule
\end{tabular}
\end{table}

\begin{table}[!ht]
\centering
\caption{AFHQV264 FID with analytical teleportation}
\label{tab:afhqv264_fid}
\begin{tabular}{rrrr}
\toprule
 Nskip &   NFE &   time/noise scale &    FID \\
\midrule
     0 &    79 &             80.0 &  2.043 \\
     2 &    75 &             60.1 &  2.029 \\
     4 &    71 &             44.6 &  2.016 \\
     6 &    67 &             32.7 &  2.003 \\
     8 &    63 &             23.6 &  2.005 \\
    10 &    59 &             16.8 &  2.026 \\
    12 &    55 &             11.7 &  2.102 \\
    14 &    51 &              8.0 &  2.359 \\
    16 &    47 &              5.4 &  3.206 \\
    20 &    39 &              2.2 & 14.442 \\
\bottomrule
\end{tabular}
\end{table}

\begin{table}[!ht]
\centering
\caption{CIFAR10 FID with analytical teleportation}
\label{tab:cifar10_fid}
\begin{tabular}{rrrr}
\toprule
 Nskip &   NFE &   time/noise scale &     FID \\
\midrule
     0 &    35 &             80.0 &   1.958 \\
     1 &    33 &             57.6 &   1.955 \\
     2 &    31 &             40.8 &   1.949 \\
     3 &    29 &             28.4 &   1.940 \\
     4 &    27 &             19.4 &   1.932 \\
     5 &    25 &             12.9 &   1.934 \\
     6 &    23 &              8.4 &   1.963 \\
     7 &    21 &              5.3 &   2.123 \\
     8 &    19 &              3.3 &   3.213 \\
    10 &    15 &              1.1 &  23.947 \\
    12 &    11 &              0.3 & 109.178 \\
\bottomrule
\end{tabular}
\end{table}

\subsection{Additional Validation experiments with Variance Preserving Diffusion }\label{apd:validation_ddpm}
Here we compared the Gaussian solutions for variance-preserving ODEs (Sec.\ref{apd:vp-ode-gaussian-solu}), with the actual diffusion trajectories sampled by DDIM sampler. We used pre-trained DDPM / DDIM models on MNIST, CIFAR10, LSUN church, CelebA 256. The dataset details are listed below (Tab. \ref{tab:ddpm-models-list}). 

\begin{table}[!ht]
\begin{center}
\caption{\textbf{Diffusion models used for numerical experiments}. \\$\dag$: The MNIST diffusion model uses the upsampled $3\times 32\times 32$ RGB pixel space as sample space, while the original MNIST data set consists of $28\times 28$ single channel black and white images. Thus the effective dimensionality of these images is around $784$. }
\label{tab:ddpm-models-list}
\begin{sc}
\begin{tabular}{lcc}
\toprule
Data set & Hugging Face model\_id & Dimensionality\\
\midrule
MNIST  & \text{dimpo/ddpm-mnist} & 3x32x32=3072 $\dag$ \\
CIFAR-10  & \text{google/ddpm-cifar10-32} & 3x32x32=3072 \\
Lsun-Church  & \text{google/ddpm-church-256} & 3x256x256 = 196,608\\
CelebA-HQ  & \text{google/ddpm-celebahq-256} & 3x256x256 = 196,608 \\
\bottomrule
\end{tabular}
\end{sc}
\end{center}
\vskip -0.1in
\end{table}

\begin{figure*}[!ht]
\vspace{-2pt}
\centering
\includegraphics[width=1.0\textwidth]{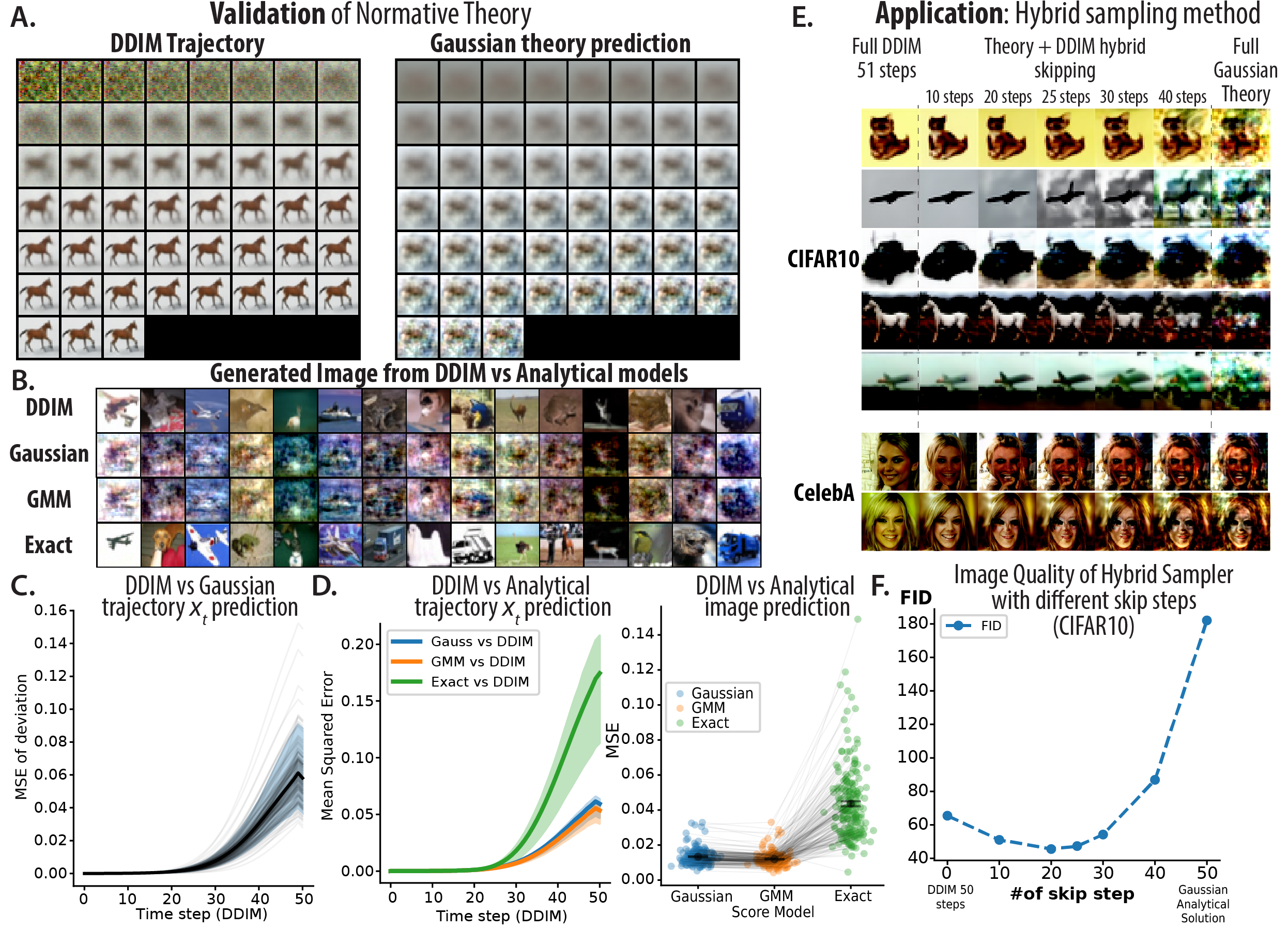}\vspace{-2pt}

\caption{\textbf{Comparing analytical solutions of sampling trajectory with DDIM diffusion model for CIFAR-10}. \textbf{A.} $\hat{\mathbf{x}}_0(\mathbf{x}_t)$ of a DDIM trajectory and the Gaussian solution with the same initial condition $\mathbf{x}_T$. \textbf{B.} Samples generated by DDIM and the analytical theories from the same initial condition. \textbf{C.} Mean squared error between the $\mathbf{x}_t$ trajectory of DDIM and Gaussian solution. \textbf{D.} Comparing the state trajectory and final sample of three normative models (Gaussian, GMM, exact) with DDIM. \textbf{E.} Hybrid sampling method combines Gaussian theory prediction with DDIM. \textbf{F.} Image quality of the hybrid method (FID score) as function of different numbers of skipped steps (see Appendix \ref{apd:fid_method}).}

\label{fig:CIFAR_theory_valid}
\vspace{-2pt}
\end{figure*}

\begin{figure}[!ht]
\vspace{-2pt}
\begin{center}
\centerline{\includegraphics[width=1.0\columnwidth]{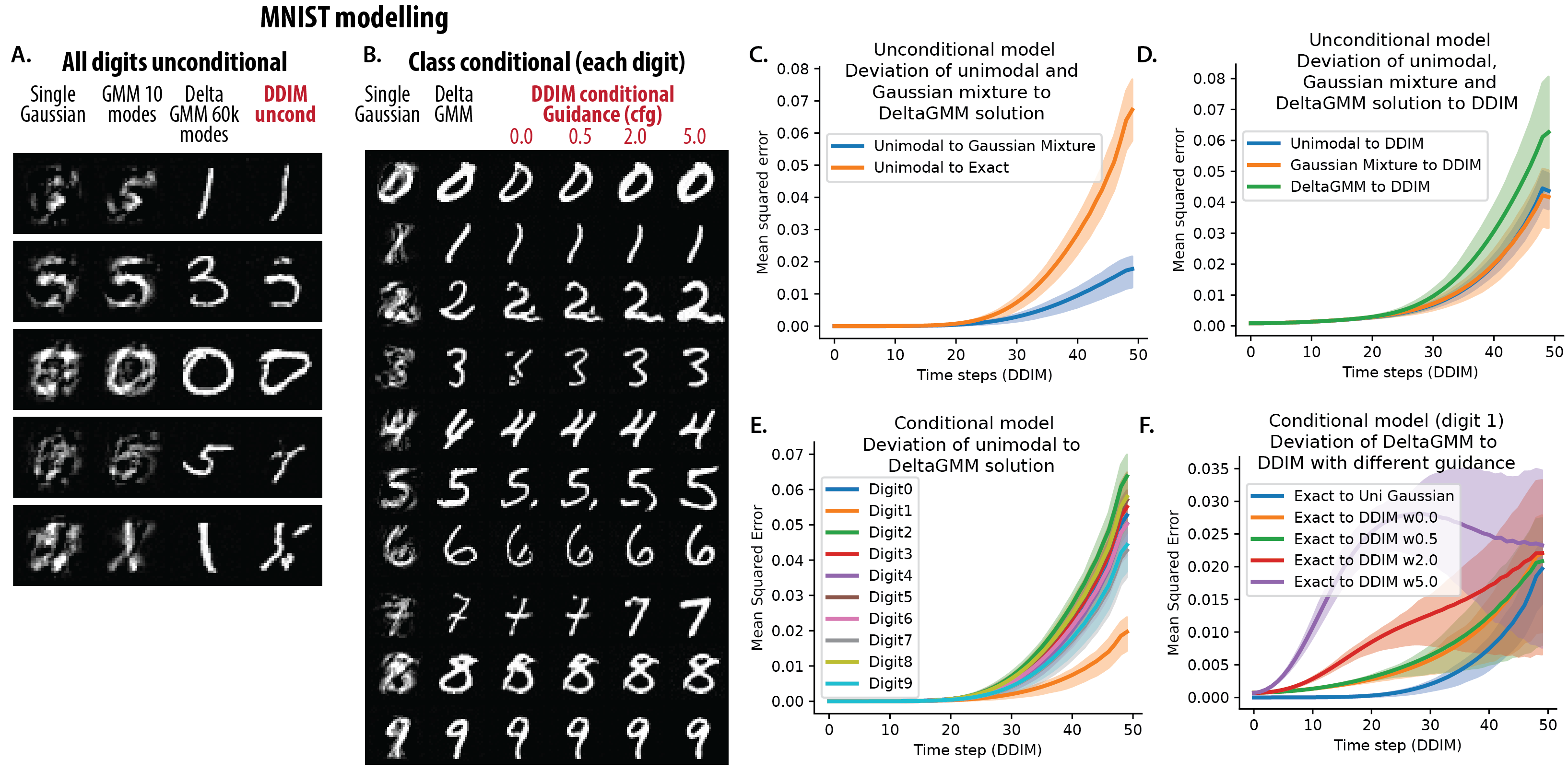}}\vspace{-10pt}
\caption{
\textbf{Comparing diffusion dynamics guided by Gaussian and Gaussian mixture score with the ones learned by a neural network (MNIST)}. 
\textbf{A.} Unconditional model. In each row, it shows the image generated from single-mode Gaussian, 10-mode Gaussian mixture, Delta GMM defined on the whole MNIST dataset, and the unconditional diffusion model. \\
\textbf{B.} Conditional model of the 10 classes. In each row, from left to right, it shows the image generated via single Gaussian, delta GMM defined on all training data of that class, and the conditional diffusion model with different classifier free guidance strength 0.0 - 5.0. \\
\textbf{C.} Unconditional model, Deviation between trajectory predicted by unimodal Gaussian theory and 10-mode GMM and exact delta GMM. It shows that Gaussian mixture the same effect on the diffusion trajectory in the first phase as a matched unimodal Gaussian.  \\
\textbf{D.} Unconditional model, Deviation between the trajectory predicted by different GMM theories and that sampled by DDIM. It shows that the trajectory sampled by DDIM is actually closer to that predicted by unimodal or 10 mode GMM, than the exact delta gmm model. This means the model didn't learn the exact score, but some coarse-grained approximation to it. \\
\textbf{E.} Conditional model, Deviation between unimodal Gaussian theory and exact delta GMM for different digits class. It shows some digits are better approximated by Gaussian than others. \\
\textbf{F.} Conditional model, Deviation between the trajectory predicted by exact delta GMM and that sampled by DDIM with different guidance scales. It shows that a larger guidance scale push the sampling result from the exact training distribution further. 
}

\label{fig:MNIST_gmm_theory_valid}
\end{center}
\end{figure}

\paragraph{Teleportation results}
For the MNIST and CIFAR-10 models, we can easily skip \textit{40\% of the initial steps} with the Gaussian solution without much of a perceptible change in the final sample (Fig.\ref{fig:CIFAR_theory_valid}E, Fig.\ref{fig:MNIST_gmm_theory_valid}D). Quantitatively, for CIFAR10 model, we found skipping up to 40\% of the initial steps can even slightly decrease the Frechet Inception Distance score (FID), and hence improve the quality of generated samples (Fig.\ref{fig:CIFAR_theory_valid}F). For models of higher resolution data sets like CelebA-HQ, we need to be more careful; skipping more than 20\% of the initial steps will induce some perceptible distortions in the generated images (Fig.\ref{fig:CIFAR_theory_valid}E bottom), which suggests that the Gaussian approximation is less effective for larger images. The reason may have to do with a low-quality covariance matrix estimate, which could arise from the number of training images being small compared to the effective dimensionality of the image manifold.

\clearpage

\section{Detailed Derivations}
\subsection{Detailed derivation to the solution of the Gaussian Diffusion} \label{apd:deriv_gaussian_model}
In this section, we derive the analytic solution $\mathbf{x}_t$ for the reverse diffusion trajectory of the Gaussian score model. We will assume that the score function corresponds to a Gaussian distribution whose mean is $\mathbf{\mu}$ and covariance matrix is $\mathbf{\Sigma}$. We will also assume, given that images are usually regarded as residing on a low-dimensional manifold within pixel space, that the rank $r$ of the covariance matrix may be less than the dimensionality $D$ of state space. Let $\mathbf{\Sigma}=\mathbf{U}\mathbf{\Lambda} \mathbf{U}^T$ be the eigendecomposition or compact SVD of the covariance matrix, where $\mathbf{U}$ is a $D \times r$ semi-orthogonal matrix whose columns are normalized (i.e. $\mathbf{U}^T\mathbf{U}=\mathbf{I}_r$), and $\mathbf{\Lambda}$ is the $r \times r$ diagonal eigenvalue matrix. Denote the $k$th column of $\mathbf{U}$ by $\mathbf{u}_k$ and the $k$th diagonal element of $\mathbf{\Lambda}$ by $\lambda_k$. 

The probability flow ODE used for the EDM model reads, 
\begin{align}
    d\mathbf{x} &=-\sigma (\sigma^2 I + \Sigma)^{-1}(\mu-\mathbf{x})d\sigma\\
    d\mathbf{x} &=-\frac{1}{\sigma}(I- \mathbf{U}\tilde \Lambda_\sigma  \mathbf{U}^T) (\mathbf{\mu}-\mathbf{x})d\sigma\;,\;\;   \\
    \tilde \Lambda_\sigma &= diag\big[\frac{\lambda_k}{\lambda_k+\sigma^2}\big]
\end{align}
We can see it's a linear ODE with respect to $\mathbf{x}-\mu$, with disentangled dynamics along each eigen mode of $\mathbf{u}_k$. Choosing our dynamic variable to be projection coefficient on each axes $c_k(\sigma)=\mathbf{u}_k^T(\mathbf{x}-\mu)$. Then its dynamics could be written as
\begin{align}
    d c_k(\sigma) = \frac{\sigma}{\lambda_k+\sigma^2}c_k(\sigma)d\sigma
\end{align}
Integrating this, we get 

\begin{align}
    \frac{d c_k(\sigma)}{c_k(\sigma)} &= d\log\sqrt{\lambda_k+\sigma^2}\\
    d\log c_k(\sigma) &= d\log\sqrt{\lambda_k+\sigma^2}\\
    c_k(\sigma) &= \sqrt{\lambda_k+\sigma^2} K
\end{align}
With integral constant $K$. Using the initial condition $c_k(T)=\mathbf{u}_k^T(\mathbf{x}_T-\mu)$, then 
\begin{align}
    \frac{c_k(\sigma)}{c_k(T)}=\sqrt{\frac{\lambda_k+\sigma^2}{\lambda_k+\sigma_T^2}}=:\psi(\lambda_k,\sigma)
\end{align}
Thus, we can arrive at the solution Eq.\ref{eq:xt_solu_psi_def}.
\begin{align}
    \mathbf{x}_\sigma-\mu &= \sum_k c_k(\sigma)\mathbf{u}_k\\
    \mathbf{x}_\sigma &= \mu + \sum_k \psi(\lambda_k,\sigma) c_k(T)\mathbf{u}_k\\
    \mathbf{x}_\sigma &= \mu + \sum_k \sqrt{\frac{\lambda_k+\sigma^2}{\lambda_k+\sigma_T^2}} \mathbf{u}_k \mathbf{u}_k^T(\mathbf{x}_T-\mu)
\end{align}
When the $\mathbf{U}$ is rank deficient, we will have the off-manifold term as in Eq. \ref{eq:xt_solu_psi_def}.

For the denoising outcome, we have Eq. \ref{eq:denoisScoreMatch}, 
\begin{align}
    D(\mathbf{x},\sigma) &= \mathbf{x} +\sigma^2 \nabla \log p (\mathbf{x},\sigma)\\
    &=\mathbf{x} + (I- \mathbf{U}\tilde \Lambda_\sigma  \mathbf{U}^T) (\mathbf{\mu}-\mathbf{x})\\
    &=\mathbf{\mu}+\mathbf{U}\tilde \Lambda_\sigma  \mathbf{U}^T(\mathbf{x} - \mathbf{\mu})
\end{align}
Using the solution above, we have
\begin{align}
    D(\mathbf{x}_\sigma,\sigma) &= \mathbf{\mu}+ \sum_k \frac{\lambda_k}{\sqrt{(\lambda_k+\sigma^2)(\lambda_k+\sigma_T^2)}} \mathbf{u}_k \mathbf{u}_k^T(\mathbf{x}_T-\mu)
\end{align}
If we define 
\begin{align}
    \xi(\lambda_k,\sigma) := \frac{\lambda_k}{\sqrt{(\lambda_k+\sigma^2)(\lambda_k+\sigma_T^2)}}
\end{align}
and use the definition of $c_k(T)$, then we get the trajectory for denoiser image as Eq.\ref{eq:denoiser_gauss_solu}.
\begin{align}
    D(\mathbf{x}_\sigma,\sigma) &= \mathbf{\mu}+ \sum_k \xi(\lambda_k,\sigma)c_k(T)\mathbf{u}_k
\end{align}

\subsection{Detailed derivation to the solution of the Gaussian Diffusion with VP-SDE} \label{apd:vp-ode-gaussian-solu}

As an extension, we present the solution for the VP-SDE forward process, which has a data scaling term $\alpha_t$ for $\mathbf{x}$. As you will see, the solutions are largely the same, just with an additional $\alpha_t$ term.

\paragraph{Forward/reverse process.} We consider forward processes defined by the stochastic differential equation (SDE) 
\begin{equation} \label{eq:forward_special}
\dot{\mathbf{x}} = -\beta(t)\mathbf{x}+g(t)\mathbf{\eta}(t)
\end{equation}
where $\beta(t)$ controls the decay of signal, $g(t)$ is a time-dependent noise amplitude, $\mathbf{\eta}(t)$ is a vector of independent Gaussian white noise terms, and time runs from $t = 0$ to $T$. Its reverse process is
\begin{equation} \label{eq:rev_special}
    \dot{\mathbf{x}} = -\beta(t)\mathbf{x}- g(t)^2 \mathbf{s}(\mathbf{x}, t) + g(t)\mathbf{\eta}(t)
\end{equation}
where $\mathbf{s}(\mathbf{x}, t) := \nabla_{\mathbf{x}} \log p(\mathbf{x}, t)$ is the score function, and where we use the standard convention that time runs \textit{backward}, i.e. from $t = T$ to $0$. In this section, we focus on one popular forward process: the variance-preserving SDE, which enforces the constraint $\beta(t)=\frac 12 g^2(t)$. The marginal probabilities of this process are
\begin{align}
p(\mathbf{x}_t | \mathbf{x}_0) = \mathcal{N}(\mathbf{x}_t | \alpha_t \mathbf{x}_0, \sigma_t^2 \mathbf{I}) \hspace{0.3in}\\
\alpha_t := e^{- \int_0^t \beta(t') dt'} \hspace{0.3in} \sigma_t^2 := 1 - e^{ - 2 \int_0^t \beta(t') dt' }
\end{align}
where $\alpha_t$ and $\sigma_t$ represent the signal and noise scale, satisfying $\alpha_t^2 + \sigma_t^2 = 1$. Normally, as $t$ goes from $0\to T$, signal scale $\alpha_t$ monotonically decreases from $1\to 0$ and $\sigma$ increases from $0\to 1$. 
Note that there exists a deterministic \textit{probability flow ODE} with the same marginal probabilities \cite{song2021scorebased} as the reverse SDE:
\begin{equation} \label{eq:rev_flow}
        \dot{\mathbf{x}} = -\beta(t)\mathbf{x} - \frac{1}{2}g(t)^2 \mathbf{s}(\mathbf{x},t)
\end{equation}
where time again runs backward from $t = 1$ to $t = 0$. In practice, instead of the SDE (Eq.\ref{eq:forward_special}), this deterministic process is often used to sample from the distribution \cite{karras2022elucidatingDesignSp}. The behavior of the probability flow ODE will be our main focus. 

\paragraph{Gaussian Score model}
For a Gaussian score model, the probability flow ODE that reverses a VP-SDE forward process \cite{song2021scorebased} is
\begin{equation}
\dot{\mathbf{x}}=-\beta(t)\mathbf{x}-\frac 12g^2(t) (\sigma_{t}^2 \mathbf{I}+\alpha_{t}^2\mathbf{\Sigma})^{-1}(\alpha_{t}\mathbf{\mu}-\mathbf{x}) \ .    
\end{equation}
Using the decomposition of $\mathbf{\Sigma}$ described above \cite{woodbury1950inverting}, 
\begin{align}
\dot{\mathbf{x}}&=-\beta(t)\mathbf{x}-\frac 12g^2(t) \frac{1}{\sigma_t^2} (\mathbf{I}-\mathbf{U}\tilde{\mathbf{\Lambda}}_t \mathbf{U}^T)(\alpha_{t}\mathbf{\mu}-\mathbf{x}) 
\end{align}
where $\tilde{\mathbf{\Lambda}}_t$ is defined to be the time-dependent diagonal matrix 
\begin{align}
\tilde{\mathbf{\Lambda}}_t &=\text{diag}\left[ \frac{\alpha_t^2\lambda_k}{\alpha_t^2\lambda_k + \sigma_t^2} \right] \ .
\end{align}
Consider the dynamics of the quantity $\mathbf{x}_t-\alpha_t \mathbf{\mu}$. Using the relationship between $\beta_t$ and $\alpha_t$, we have
\begin{align}
    \frac{d}{dt} ( \mathbf{x}_t-\alpha_t \mathbf{\mu} ) &=\dot{\mathbf{x}}_t -\mathbf{\mu} \dot \alpha_t \\
    &=\dot{\mathbf{x}}_t +\beta_t\alpha_t \mathbf{\mu} \\
    &=\beta_t(\alpha_t\mathbf{\mu}-\mathbf{x})-\frac 12g^2(t) \frac{1}{\sigma_t^2} (\mathbf{I}-\mathbf{U}\tilde{\mathbf{\Lambda}}_t \mathbf{U}^T)(\alpha_{t}\mathbf{\mu}-\mathbf{x})\\
    &= \left[ \frac 12g^2(t) \frac{1}{\sigma_t^2} (\mathbf{I}-\mathbf{U}\tilde{\mathbf{\Lambda}}_t \mathbf{U}^T)-\beta_t \mathbf{I} \right](\mathbf{x} - \alpha_{t}\mathbf{\mu}) \ .
\end{align}
Assuming that the forward process is a variance-preserving SDE, then $\beta_t=\frac12 g^2(t)$, which implies $\alpha_t^2=1-\sigma_t^2$. Using this, we obtain
\begin{align}
    \frac{d}{dt}(\mathbf{x}_t-\alpha_t \mathbf{\mu}) 
    &=\beta_t \left[ \frac{1}{\sigma_t^2} (\mathbf{I}-\mathbf{U}\tilde{\mathbf{\Lambda}}_t \mathbf{U}^T)-\mathbf{I} \right](\mathbf{x} - \alpha_{t}\mathbf{\mu})\\
    &=\beta_t \left[ (\frac{1}{\sigma_t^2} - 1)\mathbf{I}-\frac{1}{\sigma_t^2}\mathbf{U}\tilde{\mathbf{\Lambda}}_t \mathbf{U}^T) \right](\mathbf{x} - \alpha_{t}\mathbf{\mu}) \ .
\end{align}
Define the variable $\mathbf{y}_t:= \mathbf{x}_t-\alpha_t \mathbf{\mu} $. We have just shown that its dynamics are fairly `nice', in the sense that the above equation is well-behaved separable linear ODE. As we are about to show, it is exactly solvable.

Write $\mathbf{y}_t$ in terms of the orthonormal columns of $\mathbf{U}$ and a component that lies entirely in the orthogonal space $\mathbf{U}^\perp$:
\begin{equation}
    \mathbf{y}_t = \mathbf{y}^{\perp}(t)+\sum_{k=1}^r c_k(t) \mathbf{u}_k \ , \ \mathbf{y}^{\perp}(t) \in \mathbf{U}^\perp \ .
\end{equation}
The dynamics of the coefficient $c_k(t)$ attached to the eigenvector $\mathbf{u}_k$ are 
\begin{align}
    \dot c_k(t)=\frac{d}{dt}(\mathbf{u}_k^T \mathbf{y}_t)&=\beta_t \left[ (\frac{1}{\sigma_t^2} - 1)-\frac{1}{\sigma_t^2} \frac{\alpha_t^2\lambda_k}{\alpha_t^2\lambda_k + \sigma_t^2} \right](\mathbf{u}_k^T \mathbf{y}_t)\\
    &=\frac{\beta_t}{\sigma_t^2} \left( 1-\sigma_t^2-\frac{\alpha_t^2\lambda_k}{\alpha_t^2\lambda_k + \sigma_t^2} \right)c_k(t)\\
    &=\frac{\beta_t\alpha_t^2}{\sigma_t^2}\left( 1-\frac{\lambda_k}{\alpha_t^2\lambda_k + \sigma_t^2} \right) c_k(t) \ .
\end{align}
Using the constraint that $\alpha_t^2+\sigma_t^2=1$, this becomes
\begin{align}
\dot c_k(t)&=\frac{\beta_t\alpha_t^2(1-\lambda_k)}{\alpha_t^2\lambda_k + \sigma_t^2}c_k(t) \ .
\end{align}
For the orthogonal space component $\mathbf{y}^{\perp}(t)$, it will stay in the orthogonal space $\mathbf{U}^{\perp}$, and more specifically the 1D space spanned by the initial $\mathbf{y}^{\perp}(t)$---so, when going backward in time, its dynamics is simply a downscaling of $\mathbf{y}^{\perp}(T)$.
\begin{align}
    \dot{\mathbf{y}}^{\perp}(t)&=\beta_t \left( \frac{1}{\sigma_t^2}-1 \right)\mathbf{y}^{\perp}(t) =\beta_t\frac{1-\sigma_t^2}{\sigma_t^2}\mathbf{y}^{\perp}(t)
    =\frac{\beta_t\alpha_t^2}{\sigma_t^2}\mathbf{y}^{\perp}(t) \ .
\end{align}
Combining these two results and solving the ODEs in the usual way, we have the trajectory solution
\begin{align}
    \mathbf{y}_t &= d(t) \mathbf{y}^{\perp}(T)+\sum_{k=1}^r c_k(t)\mathbf{u}_k\\
    d(t)&=\exp\left(\int_T^t d\tau \frac{\beta_\tau\alpha_\tau^2}{\sigma_\tau^2}\right)\\
    c_k(t)&=c_k(T)\exp\left(\int_T^t d\tau \frac{\beta_\tau\alpha_\tau^2(1-\lambda_k)}{\alpha_\tau^2\lambda_k + \sigma_\tau^2}\right) \ .
\end{align}
The initial conditions are
\begin{align}
    c_k(T)&=\mathbf{u}_k^T \mathbf{y}_T\\
    \mathbf{y}^{\perp}(T)&=\mathbf{y}_T-\sum_{k=1}^r c_k(T)\mathbf{u}_k \ , \ \mathbf{y}^{\perp}(T)\in \mathbf{U}^\perp \ .
\end{align}
To solve the ODEs, it is helpful to use a particular reparameterization of time. In particular, consider a reparameterization in terms of $\alpha_t$ using the relationship $-\beta_t \alpha_t dt=d\alpha_t$. The integral we must do is
\begin{align}
    \int_T^t d\tau \frac{\beta_\tau\alpha_\tau^2(1-\lambda_k)}{\alpha_\tau^2\lambda_k + \sigma_\tau^2}
    &=\int_T^t d\tau \frac{\beta_\tau\alpha_\tau^2(1-\lambda_k)}{1+\alpha_\tau^2(\lambda_k -1)}\\
    &=\int_{\alpha_T}^{\alpha_t} d\alpha_\tau \frac{\alpha_\tau(\lambda_k-1)}{1+\alpha_\tau^2(\lambda_k -1)}\\
    &=\frac 12 \log(1+\alpha_\tau^2(\lambda_k -1))\Bigr|^{\alpha_t}_{\alpha_T}\\
    &=\frac 12 \log\left(\frac{1+(\lambda_k -1)\alpha_t^2}{1+(\lambda_k -1)\alpha_T^2}\right) \ .
\end{align}
Note that taking $\lambda_k=0$ gives us the solution to dynamics in the directions orthogonal to the manifold. We have 
\begin{align}
    c_k(t)&=c_k(T)\sqrt{\frac{1+(\lambda_k-1)\alpha_t^2}{1+(\lambda_k-1)\alpha_T^2}}=c_k(T)\sqrt{\frac{\sigma_t^2+\lambda_k\alpha_t^2}{\sigma_T^2+\lambda_k\alpha_T^2}}\\
    d(t)&=\sqrt{\frac{1-\alpha_t^2}{1-\alpha_T^2}}=\frac{\sigma_t}{\sigma_T} \ .
\end{align}
The time derivatives of these coefficients are
\begin{align}
\dot{c}_k(t)&=c_k(T)\frac{-(\lambda_k-1)\alpha_t^2\beta_t}{\sqrt{(1+(\lambda_k-1)\alpha_T^2)({1+(\lambda_k-1)\alpha_t^2)}}}\\
\dot{d}(t)&=\frac{\alpha_t^2\beta_t}{\sqrt{(1-\alpha_T^2)(1-\alpha_t^2)}} \ .
\end{align}
Finally, we can write out the explicit solution for the trajectory $\mathbf{x}_t$: 
\begin{align}
    \mathbf{x}_t = \alpha_t \mathbf{\mu} + d(t) \mathbf{y}^{\perp}(T)+\sum_{k=1}^r c_k(t)\mathbf{u}_k \ .
\end{align}
The solution is a sum of on- and off-manifold components:
\begin{equation}  \label{eq:xt_solu_psi_def_apd}
\begin{split}
\mathbf{x}_t = \alpha_t \mathbf{\mu} + \frac{\sigma_t}{\sigma_T} \ \mathbf{y}^{\perp}_T + \sum_{k=1}^r \psi(t, \lambda_k) c_k(T) \mathbf{u}_k  \hspace{0.5in}
\psi(t, \lambda_k)= \sqrt{  \frac{\sigma_t^2 + \lambda_k \alpha_t^2}{\sigma_T^2 + \lambda_k \alpha_T^2} }  \\
\mathbf{y}^{\perp}_T=\ (\mathbf{I}-\mathbf{U}^T\mathbf{U})(\mathbf{x}_T-\alpha_T \mathbf{\mu})
\hspace{0.5in}
c_k(T) = \ \mathbf{u}_k^T(\mathbf{x}_T-\alpha_T \mathbf{\mu}) \ .
\end{split}
\end{equation}  
The initial condition $\mathbf{x}_T$ can be decomposed into contributions along each principal direction, an off-manifold contribution, and a $\alpha_T \mathbf{\mu}$ contribution. There are three terms in the dynamics for these three parts: 1) the scaling up of the distribution mean; 2) the scaling down of the off-manifold component $\mathbf{y}^{\perp}_T$, proportional to the noise scale $\psi(t,0)=\frac{\sigma_t}{\sigma_T}$; and 3) the on-manifold movement along each eigenvector governed by $\psi(t,\lambda_k)$. 


We also now have the analytical solution for the projected outcome or the ideal denoiser, which is defined as:
\begin{equation}
    D(\mathbf{x}_t,\sigma_t) := \frac{\mathbf{x}_t+\sigma_t^2 \mathbf{s}(\mathbf{x}_t,t)}{\alpha_t} 
\end{equation}

\begin{align}
    D(\mathbf{x}_t,\sigma_t)-\mathbf{\mu}
    &=\frac{1}{\alpha_t}\mathbf{U}\tilde{\mathbf{\Lambda}}_t \mathbf{U}^T(\mathbf{x}_t-\alpha_t \mathbf{\mu})\\
    &=\sum_{k=1}^rc_k(t) \frac{\alpha_t\lambda_k}{\alpha_t^2\lambda_k + \sigma_t^2} \mathbf{u}_k \notag \\
    &=\sum_{k=1}^rc_k(T) \frac{\alpha_t\lambda_k}{\sqrt{(\alpha_t^2\lambda_k + \sigma_t^2)(\alpha_T^2\lambda_k + \sigma_T^2)}} \mathbf{u}_k \ . \notag
\end{align}
Thus,
\begin{align}
    D(\mathbf{x}_t,\sigma_t) = \mathbf{\mu} +
    \sum_{k=1}^rc_k(T) \frac{\alpha_t\lambda_k}{\sqrt{(\alpha_t^2\lambda_k + \sigma_t^2)(\alpha_T^2\lambda_k + \sigma_T^2)}} \mathbf{u}_k 
\end{align}

\newpage
\subsection{Score function of general Gaussian mixture models}\label{apd:gmm_score}
Let 
\begin{equation}
q(\mathbf{x}) = \sum_i \pi_i \mathcal N(\mathbf{x};\mathbf{\mu}_i,\mathbf{\Sigma}_i)
\end{equation}
be a Gaussian mixture distribution, where the $\pi_i$ are mixture weights, $\mathbf{\mu}_i$ is the $i$-th mean, and $\mathbf{\Sigma}_i$ is the $i$-th covariance matrix. The score function for this distribution is
\begin{equation}
\begin{split}
  \nabla_{\mathbf{x}} \log q(\mathbf{x}) = &\frac{\sum_i \pi_i \nabla_{\mathbf{x}} \mathcal N(\mathbf{x};\mathbf{\mu}_i,\mathbf{\Sigma}_i)}{q(\mathbf{x})}\\
    =&\sum_i -\Sigma_i^{-1}(\mathbf{x}-\mathbf{\mu}_i)\frac{\pi_i \mathcal N(\mathbf{x};\mathbf{\mu}_i,\mathbf{\Sigma}_i)}{q(\mathbf{x})}\\
    =&\sum_i \frac{\pi_i \mathcal N(\mathbf{x};\mathbf{\mu}_i,\mathbf{\Sigma}_i)}{q(\mathbf{x})}\nabla_\mathbf{x} \log \mathcal N(\mathbf{x};\mathbf{\mu}_i,\mathbf{\Sigma}_i) \\
    =&\sum_i w_i(\mathbf{x})\nabla_\mathbf{x} \log \mathcal N(\mathbf{x};\mathbf{\mu}_i,\mathbf{\Sigma}_i) 
\end{split}
\end{equation}
where we have defined the mixing weights
\begin{align}
    w_i(\mathbf{x}):=\frac{\pi_i \mathcal N(\mathbf{x};\mathbf{\mu}_i,\mathbf{\Sigma}_i)}{q(\mathbf{x})}  \ .
\end{align}
Thus, the score of the Gaussian mixture is a weighted mixture of the score fields of each of the individual Gaussians. 

In the context of diffusion, we are interested in the \textit{time-dependent} score function. Given a Gaussian mixture initial condition, the end result of the VP-SDE forward process will also be a Gaussian mixture:
\begin{align}
    p_t(\mathbf{x}) = \sum_i \pi_i \mathcal N(\mathbf{x}; \mathbf{\mu}_i,\sigma_t^2 \mathbf{I}+ \mathbf{\Sigma}_i) \ .
\end{align}
The corresponding time-dependent score is
\begin{equation}
\begin{split}
\mathbf{s}(\mathbf{x},t)&=\nabla_{\mathbf{x}} \log p_t(\mathbf{x})\\
=&\sum_i -(\sigma_t^2 \mathbf{I}+ \mathbf{\Sigma}_i)^{-1}(\mathbf{x}- \mathbf{\mu}_i)\frac{\pi_i \mathcal N(\mathbf{x}; \mathbf{\mu}_i,\sigma_t^2 \mathbf{I}+\mathbf{\Sigma}_i)}{p_t(\mathbf{x})}\\
=&\sum_i -(\sigma_t^2 \mathbf{I}+ \mathbf{\Sigma}_i)^{-1}(\mathbf{x}-\mathbf{\mu}_i)w_i(\mathbf{x}, t) \ .
\end{split}
\end{equation}
Note that we have a formula for $(\sigma_t^2 \mathbf{I}+ \mathbf{\Sigma}_i)^{-1}$ (derived using the Woodbury matrix inversion identity; see Eq.\ref{eq:gauss_score}) in terms of the (compact) SVD of $\mathbf{\Sigma}_i$. We can use it to write
\begin{align}\label{eq:efficient_gmm_score}
\mathbf{s}(\mathbf{x},t)=\frac{1}{\sigma_t^2} \sum_i -(\mathbf{I}-\mathbf{U}_i\tilde{\mathbf{\Lambda}_i}_t \mathbf{U}_i^T)(\mathbf{x}- \mathbf{\mu}_i) w_i(\mathbf{x}, t)
\end{align}
where
\begin{equation}
\mathbf{\Sigma}_i=\mathbf{U}_i\mathbf{\Lambda}_i\mathbf{U}_i^T \hspace{1in} \tilde{\mathbf{\Lambda}}_t :=\text{diag}\left[\frac{\lambda_k}{\lambda_k + \sigma_t^2} \right] \ .
\end{equation}
This representation of the score function is numerically convenient, since (once the SVDs of each covariance matrix have been obtained), it can be evaluated using a relatively small number of matrix multiplications, which are cheaper than the covariance matrix inversions that a naive implementation of the Gaussian mixture score function would require.


We used this formula (Eq.\ref{eq:efficient_gmm_score}) and an off-the-shelf ODE solver to simulate the reverse diffusion trajectory of a 10-mode Gaussian mixture score model (Fig. \ref{fig:CIFAR_theory_valid}).

\subsection{Score function of Gaussian mixture with identical and isotropic covariance}
Assume each Gaussian mode has covariance $\mathbf{\Sigma}_i=\sigma^2 \mathbf{I}$ and that every mixture weight is the same (i.e. $\pi_i=\pi_j,\forall i,j$). Then the score for this kind of specific Gaussian mixture is 
\begin{equation}\label{eq:gmm_isotrop}
\begin{split}
\nabla_{\mathbf{x}} \log q(\mathbf{x}) = &\frac{\sum_i \pi_i \nabla_{\mathbf{x}} \mathcal N(\mathbf{x};\mathbf{\mu}_i,\sigma^2 \mathbf{I})}{\sum_i \pi_i \mathcal N(\mathbf{x};\mathbf{\mu}_i,\sigma^2 \mathbf{I})}\\
    =&\frac{\sum_i -\frac{1}{\sigma^2}(\mathbf x-\mathbf{\mu}_i) \exp \big(-\frac{1}{2\sigma^2}\|\mathbf x - \mathbf{\mu}_i\|_2^2\big)}{\sum_i\exp \big(-\frac{1}{2\sigma^2}\|\mathbf x - \mathbf{\mu}_i\|_2^2\big)}\\
    =&\frac{1}{\sigma^2}\sum_i w_i(\mathbf{x})(\mathbf{\mu}_i - \mathbf x) \ ,
\end{split}
\end{equation}
where the weight $w_i(\mathbf{x})$ is a softmax of the negative squared distance to all the means, with $\sigma^2$ functioning as a temperature parameter:
\begin{align}\label{eq:gmm_w_softmax}
    w_i(\mathbf{x}) &= Softmax(\big\{-\frac{1}{2\sigma^2}\|\mathbf x - \mathbf{\mu}_i\|_2^2\big\}) =\frac{\exp \big(-\frac{1}{2\sigma^2}\|\mathbf x - \mathbf{\mu}_i\|_2^2\big)}{\sum_i\exp \big(-\frac{1}{2\sigma^2}\|\mathbf x - \mathbf{\mu}_i\|_2^2\big)} \ .
\end{align}
Since the weights $w_i(\mathbf{x})$ sum to $1$, we can also write the score function in the suggestive form
\begin{align}
\nabla_{\mathbf{x}} \log q(\mathbf{x}) = \frac{(\sum_i w_i(\mathbf{x}) \mathbf{\mu}_i )-\mathbf{x}}{\sigma^2} \ .
\end{align}
This has a form analogous to the score of a single Gaussian mode---but instead of $\mathbf{x}$ being `attracted' towards a single mean $\mathbf{\mu}$, it is attracted towards a weighted combination of all of the means, with modes closer to the state $\mathbf{x}$ being more highly weighted.


\subsection{Score of exact (delta mixture) score model}

A particularly interesting special case of the Gaussian mixture model is the delta mixture model used in the main text, whose components are vanishing-width Gaussians centered on the training images. In particular, consider a data set $\{\mathbf{y}_i\}$ with $i=1,...,N$, so that the starting distribution is 
\begin{equation}
p(\mathbf{x})=\frac{1}{N} \sum_i\delta (\mathbf{x}-\mathbf{y}_i) \ .
\end{equation} 
At time $t$, the marginal distribution will be a Gaussian mixture 
\begin{align}
    p_t(\mathbf{x}_t)=\frac{1}{N} \sum_i \mathcal{N}(\mathbf{x}_t;\mathbf{y}_i,\sigma_t^2 \mathbf{I}) \ .
\end{align}
Then using the Eq. \ref{eq:gmm_isotrop} above we have 
\begin{equation}
\begin{split}
    s(\mathbf x_t,t)=\nabla \log p_t(\mathbf x_t) &= \frac{1}{\sigma_t^2} \sum_i w_i(\mathbf x_t)(\mathbf{y}_i - \mathbf x_t)\\
    &= \frac{1}{\sigma_t^2} \left[- \mathbf x_t +  \sum_i w_i(\mathbf x_t)\mathbf{y}_i \right]\\
    &= \frac{1}{\sigma_t^2} \left[- \mathbf x_t +  \sum_i Softmax(\big\{-\frac{1}{2\sigma_t^2}\|\mathbf{y}_i - \mathbf x_t\|^2\big\}) \mathbf{y}_i \right] \ .
\end{split}
\end{equation}
The endpoint estimate of the distribution is 
\begin{equation}
\begin{split}
    \hat{\mathbf x}_0(\mathbf x_t)&=\frac{\mathbf{x}_t+\sigma_t^2 \nabla \log p(\mathbf x_t)}{} \\
    &=\sum_i Softmax(\big\{-\frac{1}{2\sigma_t^2}\|\mathbf{y}_i - \mathbf x_t\|^2\big\}) \mathbf{y}_i\\
    &=\sum_i w_i(\mathbf x_t)\mathbf{y}_i \ .
\end{split}
\end{equation}
Thus, the endpoint estimate is a weighted average of training data, with the softmax of negative squared distance as weights and $\sigma_t^2$ as a temperature parameter.

\newpage
\subsection{Arguments of the approximation of score field of Gaussian and point cloud}\label{apd:gauss_pointcloud_equiv}
In this section, we will argue from a theoretical perspective, the score field of a bounded point cloud is equivalent to a Gaussian with matching mean and covariance, when the noise level is high and when the query point is far from the bounded point cloud. 
We are going to use a technique inspired by multi-pole expansion in electrodynamics \citep{griffiths2005introduction}. 

\paragraph{Set up} Assume we have a set of data points $\{\mathbf{y}_i\},i=1...N$, their mean and covariance are defined as 
\begin{align}
    \mu &= \frac1N \sum_i \mathbf{y}_i\\
    \Sigma &= \frac1N \sum_i \mathbf{y}_i\mathbf{y}_i^T -\mu\mu^T
\end{align}
these are the first moment and the second central moment of the distribution. The Gaussian distribution with matching first two moments is $\mathcal{N}(\mu,\Sigma)$. We are going to show that when the noise level $\sigma$ and $\|\mathbf{x}-\mu\|$ are both much larger than the standard deviation of the distribution, the score at noise level $\sigma$ of the original dataset (point cloud) is equivalent to the score of this Gaussian distribution. 

\paragraph{Score expansion for Gaussian distribution} Consider the Gaussian distribution, at noise level $\sigma$, the distribution is $\mathcal{N}(\mu,\Sigma+\sigma^2 I)$. Then its score can be written as 
\begin{align}
    \log p(\mathbf{x};\sigma) &= (\sigma^2 I + \Sigma)^{-1}(\mathbf{\mu}-\mathbf{x})\\
    &=\frac{1}{\sigma^2}(I-U\tilde \Lambda_\sigma  U^T) (\mathbf{\mu}-\mathbf{x})
\end{align}
in which the 
\begin{align}
    \tilde \Lambda_\sigma &= diag\big[\frac{\lambda_k}{\lambda_k+\sigma^2}\big]\\
    & = diag\big[\frac{\lambda_k}{\sigma^2}-\frac{\lambda_k^2}{\sigma^2(\sigma^2+\lambda_k)}\big]\\
    & = diag\big[\frac{\lambda_k}{\sigma^2}-\frac{\lambda_k^2}{\sigma^4}+\frac{\lambda_k^3}{\sigma^4(\sigma^2+\lambda_k)}\big]\\
    & = \frac1{\sigma^2}\Lambda - \frac1{\sigma^4}\Lambda^2 + \frac1{\sigma^6}\Lambda^3 - \frac1{\sigma^8}\Lambda^4 + ...\\
    & = \frac1{\sigma^2}\Lambda - \Delta 
\end{align}
where the residual term $\Delta$ is the 2nd order term of $\lambda/\sigma^2$, $\Delta \sim (\lambda/\sigma^2)^2$. 

Using this expansion, the score field can be expressed as 
\begin{align}
    \log p(\mathbf{x};\sigma) &=\frac{1}{\sigma^2}(I- \frac1{\sigma^2} U\Lambda U^T + U\Delta U^T) (\mathbf{\mu}-\mathbf{x})\\
    &=\frac{1}{\sigma^2} (\mathbf{\mu}-\mathbf{x}) - \frac{1}{\sigma^4} \Sigma(\mathbf{\mu}-\mathbf{x}) + \frac{1}{\sigma^2}U\Delta U^T(\mathbf{\mu}-\mathbf{x})\\
    &=\frac{1}{\sigma^2} (\mathbf{\mu}-\mathbf{x}) - \frac{1}{\sigma^4} \Sigma(\mathbf{\mu}-\mathbf{x}) + \frac{1}{\sigma^6}U\Lambda^2U^T(\mathbf{\mu}-\mathbf{x}) - \frac{1}{\sigma^8}U\Lambda^3U^T(\mathbf{\mu}-\mathbf{x}) + ... \\
    &=\frac{1}{\sigma^2} (\mathbf{\mu}-\mathbf{x}) - \frac{1}{\sigma^4} \Sigma(\mathbf{\mu}-\mathbf{x}) + \frac{1}{\sigma^6}\Sigma^2(\mathbf{\mu}-\mathbf{x}) - \frac{1}{\sigma^8}\Sigma^3(\mathbf{\mu}-\mathbf{x}) + ... \label{eq:gauss_score_exp_series}
\end{align}

From this result, we can express the score field of a Gaussian distribution by a series of terms: the first-order term is the isotropic attraction to the mean, the second-order term is the anisotropic term conditioned by the covariance matrix; the higher-order terms are increasingly anisotropic with higher power of the covariance matrix, i.e. larger condition number. 

Note that, this series is always linear with respect to $\mathbf{\mu}-\mathbf{x}$, and it doesn't contain any quadratic term of $\mathbf{x}$. 

\paragraph{Score expansion for delta point distribution} 
Consider the data points, at noise level $\sigma$, the noised distribution is a Gaussian mixture with $N$ components, $q(\mathbf{x};\sigma)=\frac{1}{N}\sum_i\mathcal{N}(\mathbf{x};\mathbf{y}_i,\sigma^2 I)$. As we have derived above, the score of a Gaussian mixture with identical variance at each mode can be expressed as 
\begin{align}
    \nabla_\mathbf{x} \log q(\mathbf{x};\sigma) &= \frac{1}{\sigma^2}\sum_i w_i(\mathbf{x})(\mathbf{y}_i - \mathbf{x})\\
    & = \frac{1}{\sigma^2} \bigg(\sum_i w_i(\mathbf{x})\mathbf{y}_i  - \mathbf{x}\bigg)\\
    &= \frac{1}{\sigma^2}(\mathbf{\mu} - \mathbf{x}) + \frac{1}{\sigma^2}\sum_i w_i(\mathbf{x})(\mathbf{y}_i - \mathbf{\mu})
\end{align}
Where the weighting function is 
\begin{align}
    w_i(\mathbf{x}) = \frac{\exp\big( -\frac{1}{2\sigma^2}\|\mathbf x - \mathbf{y}_i\|^2_2\big)}{\sum^N_j \exp\big( -\frac{1}{2\sigma^2}\|\mathbf x - \mathbf{y}_j\|^2_2\big)}
\end{align}

Here we rewrite all the distances using the distributional mean as a reference 
\begin{align}
    \|\mathbf x - \mathbf{y}_i\|^2_2 &= \|\mathbf x - \mu + \mu - \mathbf{y}_i\|^2_2\\
    &=\|\mathbf x - \mu\|^2_2 + \|\mu - \mathbf{y}_i\|^2_2 + 2(\mathbf x - \mu)^T (\mu - \mathbf{y}_i)
\end{align}
Multiply the same term $\exp \big(\frac{1}{2\sigma^2} \|\mathbf x - \mu\|^2_2\big)$ at numerator and denominator, we get 
\begin{align}
    w_i(\mathbf{x}) = \frac{\exp\big( -\frac{1}{2\sigma^2}(\|\mu - \mathbf{y}_i\|^2_2 + 2(\mathbf x - \mu)^T (\mu - \mathbf{y}_i) ) \big)}{\sum^N_j \exp\big( -\frac{1}{2\sigma^2} ( \|\mu - \mathbf{y}_j\|^2_2 + 2(\mathbf x - \mu)^T (\mu - \mathbf{y}_j) )\big)}
\end{align}
Note the $(\mathbf x - \mu)^T (\mu - \mathbf{y}_i)$ is on the order of $\|\mathbf x - \mu\| \|\mu - \mathbf{y}_i\|\propto \sigma \sqrt{d} \sqrt{tr\Lambda }$. 
$\|\mu - \mathbf{y}_j\|^2_2$ is on the order of $tr\Lambda$. 
$\|\mathbf x - \mu\|^2$ is on the order of $d\sigma^2$. 

We can expand the exponential functions to the first-order of $\sqrt{d}\sqrt{tr\Lambda}/\sigma$
\begin{align}
    w_i(\mathbf{x}) \approx &\ \frac{1 - \frac{1}{\sigma^2}(\mathbf x - \mu)^T (\mu - \mathbf{y}_i) -\frac{1}{2\sigma^2}\|\mu - \mathbf{y}_i\|^2_2 }{N + \sum^N_j \big[ - \frac{1}{\sigma^2}(\mathbf x - \mu)^T (\mu - \mathbf{y}_j) -\frac{1}{2\sigma^2}\|\mu - \mathbf{y}_j\|^2_2 \big]}
    \\
    =&\ \frac{1 - \frac{1}{\sigma^2}(\mathbf x - \mu)^T (\mu - \mathbf{y}_i) -\frac{1}{2\sigma^2}\|\mu - \mathbf{y}_i\|^2_2 }{N + \sum^N_j \big[ -\frac{1}{2\sigma^2}\|\mu - \mathbf{y}_j\|^2_2 \big]}\\
    =&\ \frac{1 - \frac{1}{\sigma^2}(\mathbf x - \mu)^T (\mu - \mathbf{y}_i) -\frac{1}{2\sigma^2}\|\mu - \mathbf{y}_i\|^2_2 }{N - N \frac{tr\Sigma}{2\sigma^2} }
\end{align}

Here we used the relationship 
\begin{align}
    \sum^N_j (\mu - \mathbf{y}_j) = 0 \\ 
    \sum^N_j \|\mu - \mathbf{y}_j\|^2_2 = N tr\Sigma \label{eq:cov_dist_lemma}
\end{align}

Now consider $\sum_i w_i(\mathbf{x})\mathbf{y}_i$
\begin{align}
    \sum^N_i w_i(\mathbf{x})\mathbf{y}_i &\approx \frac{1}{N(1-tr\Sigma /2\sigma^2)} \sum^N_i \bigg[1 - \frac{1}{\sigma^2}(\mathbf x - \mu)^T (\mu - \mathbf{y}_i) -\frac{1}{2\sigma^2}\|\mu - \mathbf{y}_i\|^2_2 \bigg]\mathbf{y}_i \\ 
    &= \frac{1}{N(1-tr\Sigma /2\sigma^2)} \bigg[ N\mu - \frac{1}{\sigma^2} \sum^N_i (\mathbf x - \mu)^T (\mu - \mathbf{y}_i) \mathbf{y}_i -
    \frac{1}{2\sigma^2} \sum^N_i \|\mu - \mathbf{y}_i\|^2_2 \mathbf{y}_i
    \bigg] \label{eq:wi_formula2}
\end{align}
Here we used the relationship 
\begin{align}
    \sum^N_i (\mathbf x - \mu)^T (\mu - \mathbf{y}_i) \mathbf{y}_i& =  \sum^N_i (\mu - \mathbf{y}_i)^T (\mathbf x - \mu) \mathbf{y}_i \\
    & =  \sum^N_i \mathbf{y}_i (\mu - \mathbf{y}_i)^T (\mathbf x - \mu)  \\
    & =  \sum^N_i (\mathbf{y}_i -\mu) (\mu - \mathbf{y}_i)^T (\mathbf x - \mu)\\  
    & = - N \Sigma (\mathbf x - \mu)\label{eq:lemma3}
\end{align}
and
\begin{align}
    \sum^N_i \|\mu - \mathbf{y}_i\|^2_2 \mathbf{y}_i &= \sum^N_i \|\mu - \mathbf{y}_i\|^2_2 (\mathbf{y}_i - \mu) + \mu \sum^N_i \|\mu - \mathbf{y}_i\|^2_2\\
    &=\sum^N_i \|\mathbf{y}_i - \mu\|^2_2 (\mathbf{y}_i - \mu) + \mu N tr\Sigma \\
    &=N\mathbf{\gamma} + N tr\Sigma\mu  \label{eq:lemma4}
\end{align}
where $\mathbf{\gamma}:=1/N \sum^N_i \|\mathbf{y}_i - \mu\|^2_2 (\mathbf{y}_i - \mu)$ is a form of third centered moment, which measures the assymetry of the distribution with respect to the mean. 

Take these Eqs. \ref{eq:lemma3},\ref{eq:lemma4} into Eq. \ref{eq:wi_formula2}, we have 
\begin{align}
    \sum^N_i w_i(\mathbf{x})\mathbf{y}_i &\approx \frac{1}{N(1-tr\Sigma /2\sigma^2)} \bigg[ N\mu + \frac{1}{\sigma^2}N\Sigma(\mathbf{x} - \mu) -
    \frac{1}{2\sigma^2} \big(N\mathbf{\gamma} + N tr\Sigma\mu \big)  \bigg] \\
    &=\mu + \frac{1}{\sigma^2(1 - tr\Sigma/2\sigma^2)} \Sigma (\mathbf{x} - \mu) - \frac{1}{2\sigma^2 (1 - tr\Sigma/2\sigma^2) }\mathbf{\gamma}
\end{align}

Taking together, the score function reads, 
\begin{align}
    \nabla_\mathbf{x} \log q(\mathbf{x};\sigma) &= \frac{1}{\sigma^2} (\sum_i w_i(\mathbf{x})\mathbf{y}_i  - \mathbf{x})\\
    &\approx \frac{1}{\sigma^2} (\mu - \mathbf{x}) + \frac{1}{\sigma^4} \Sigma( \mathbf{x} - \mu) - \frac{1}{2\sigma^4}\gamma + ...\label{eq:pointcloud_score_exp_series1}
\end{align}

We can see the expansion of the score for point cloud (Eq. \ref{eq:pointcloud_score_exp_series1}) share the first two terms with the expansion of the Gaussian score (Eq. \ref{eq:gauss_score_exp_series}): the isotropic term of $(\mu-\mathbf{x})/\sigma^2$ and the first order anisotropic term determined by the covariance.



\paragraph{Alternative score expansion for delta point distribution} 
Another way of expansion is the following 
\begin{align}
    w_i(\mathbf{x}) = \frac{\exp\big( -\frac{1}{2\sigma^2}(\|\mu - \mathbf{y}_i\|^2_2 + 2(\mathbf x - \mu)^T (\mu - \mathbf{y}_i) ) - tr\Sigma \big)}
    {\sum^N_j \exp\big( -\frac{1}{2\sigma^2} ( \|\mu - \mathbf{y}_j\|^2_2 + 2(\mathbf x - \mu)^T (\mu - \mathbf{y}_j) - tr\Sigma )\big)}
\end{align}
Note the $(\mathbf x - \mu)^T (\mu - \mathbf{y}_i)$ is on the order of $\|\mathbf x - \mu\| \|\mu - \mathbf{y}_i\|\propto \sigma \sqrt{d} \sqrt{tr\Lambda }$. 
$\|\mu - \mathbf{y}_j\|^2_2$ is on the order of $tr\Lambda$. 
$\|\mathbf x - \mu\|^2$ is on the order of $d\sigma^2$. 

We can expand the exponential functions to the first-order of $\sqrt{d}\sqrt{tr\Lambda}/\sigma$
\begin{align}
    w_i(\mathbf{x}) \approx &\ \frac{1 - \frac{1}{\sigma^2}(\mathbf x - \mu)^T (\mu - \mathbf{y}_i) -\frac{1}{2\sigma^2}\|\mu - \mathbf{y}_i\|^2_2 + \frac{1}{2\sigma^2} tr\Sigma  }{N + \sum^N_j \big[ - \frac{1}{\sigma^2}(\mathbf x - \mu)^T (\mu - \mathbf{y}_j) -\frac{1}{2\sigma^2}\|\mu - \mathbf{y}_j\|^2_2 + \frac{1}{2\sigma^2} tr\Sigma \big]}
    \\
    =&\ \frac{1 - \frac{1}{\sigma^2}(\mathbf x - \mu)^T (\mu - \mathbf{y}_i) -\frac{1}{2\sigma^2}\|\mu - \mathbf{y}_i\|^2_2 + \frac{1}{2\sigma^2} tr\Sigma  }{N + \sum^N_j \big[ -\frac{1}{2\sigma^2}\|\mu - \mathbf{y}_j\|^2_2 + \frac{N}{2\sigma^2} tr\Sigma\big]}\\
    =&\ \frac{1 - \frac{1}{\sigma^2}(\mathbf x - \mu)^T (\mu - \mathbf{y}_i) -\frac{1}{2\sigma^2}\|\mu - \mathbf{y}_i\|^2_2 + \frac{1}{2\sigma^2} tr\Sigma  }{N }\\
    =&\frac{1}{N\sigma^2}\big[\sigma^2 - (\mathbf x - \mu)^T (\mu - \mathbf{y}_i) -\frac{1}{2}\|\mu - \mathbf{y}_i\|^2_2 + \frac{1}{2} tr\Sigma  \big]
\end{align}
Here we used the identity Eq.\ref{eq:cov_dist_lemma} to simplify the denominator.
\begin{align}
    \sum_i w_i(\mathbf{x}) (\mathbf{y}_i - \mathbf{\mu}) &\approx  \frac{1}{N\sigma^2} \big[ N \Sigma (\mathbf{x} - \mu)- \sum_i \frac12 \|\mu-\mathbf{y}_i\|^2(\mathbf{y}_i - \mathbf{\mu})\big]\\
    &=\frac{1}{\sigma^2}\Sigma (\mathbf{x} - \mu)-\frac{1}{2\sigma^2}\sum_i \|\mu-\mathbf{y}_i\|^2(\mathbf{y}_i - \mathbf{\mu})
\end{align}

Thus 
\begin{align}
    \nabla_\mathbf{x} \log q(\mathbf{x};\sigma) 
    &= \frac{1}{\sigma^2}(\mathbf{\mu} - \mathbf{x}) + \frac{1}{\sigma^2}\sum_i w_i(\mathbf{x})(\mathbf{y}_i - \mathbf{\mu})\\
    &\approx \frac{1}{\sigma^2}(\mathbf{\mu} - \mathbf{x}) + \frac{1}{\sigma^4}\Sigma (\mathbf{x} - \mu)-\frac{1}{2\sigma^4}\sum_i \|\mu-\mathbf{y}_i\|^2(\mathbf{y}_i - \mathbf{\mu}) + ...\label{eq:pointcloud_score_exp_series2}
\end{align}
Similarly, using this expansion (Eq. \ref{eq:pointcloud_score_exp_series2}), we can see the expansion of the score for point cloud share the first two terms with the expansion of the Gaussian score (Eq. \ref{eq:gauss_score_exp_series}): the isotropic term of $(\mu-\mathbf{x})/\sigma^2$ and the first order anisotropic term determined by the covariance. 

\paragraph{Validity of Assumption}
For a $d$ dimensional Gaussian distribution $\mathbf{x}\sim \mathcal{N}(0,\sigma^2 I)$, the squared norm $\|\mathbf{x}/\sigma\|^2\sim \chi^2(d)$, $\mathbb E\|\mathbf{x}\|^2=d\sigma^2$. 
As $\sum_i^N \|\mathbf{y}_i - \mathbf{\mu}\|^2 = N tr \Lambda$,
$\|\mathbf{y}_i - \mathbf{\mu}\|^2$ is on the order of $tr \Lambda = \sum_k\lambda_k$. 
For a bounded point cloud where $\|\mu\|\ll \|\mathbf{x}\|$ and $\|\mu - \mathbf{y}_i\|\ll \|\mathbf{x}\|$. Then, $\|\mathbf{x}-\mathbf{y}_i\|^2$ and $\|\mathbf{x}-\mu\|^2$ are also on the order of $d\sigma^2$. 
Further works are needed to show the more precise condition at which the Gaussian approximation holds well. 

\end{document}